\documentclass[preprint,12pt]{elsarticle}
\usepackage{amssymb}
\usepackage{amsmath}
\usepackage{algorithm}
\usepackage{algorithmic}
\usepackage{color}

\usepackage{multirow}
\usepackage{appendix}
\usepackage{graphicx}
\usepackage{pgfplots}
\usepackage{caption}
\usepackage{tikz}
\pgfplotsset{compat=1.18}
\DeclareMathOperator*{\argmin}{arg\,min}
\journal{Neurocomputing}
\usepackage{hyperref}
\newtheorem{defn}{Definition}
\newtheorem{prop}{Proposition}

\usepackage[defaultcolor=red]{changes}

\begin{document}
\begin{frontmatter}

\title{Adversarial Defenses via Vector Quantization}
\author[label1]{Zhiyi Dong}
\author[label1]{Yongyi Mao\corref{cor1}}
\affiliation[label1]{organization={School of Electrical Engineering and Computer Science, University of Ottawa},
            addressline={800 King Edward Avenue}, 
            city={Ottawa},
            postcode={K1N 6N5}, 
            state={Ontario},
            country={Canada}}
\cortext[cor1]{Corresponding author}

\begin{abstract}
Adversarial attacks pose significant challenges to the robustness of modern deep neural networks in computer vision, and defending these networks against adversarial attacks has attracted intense research efforts. Among various defense strategies, preprocessing-based defenses are practically appealing since there is no need to train the network under protection. However, such approaches typically do not achieve comparable robustness as other methods such as adversarial training \cite{fgsm,pgd,sankaranarayanan2018}.  In this paper, we propose a novel framework for preprocessing-based defenses, where a vector quantizer is used as a preprocessor. This framework, inspired by and extended from Randomized Discretization (``RandDisc" \cite{randdisc}), is theoretically principled by rate-distortion theory: indeed, RandDisc may be viewed as a scalar quantizer, and rate-distortion theory suggests that such quantization schemes are inferior to vector quantization.  
In our framework, the preprocessing vector quantizer treats the input image as a collection of patches and finds a set of ``representative patches" based on the patch distributions; each original patch is then modified according to the representative patches close to it. 

We present two lightweight defenses in this framework, referred to as patched RandDisc (pRD) and sliding-window RandDisc (swRD), where the patches are disjoint in the former and overlapping in the latter. We show that vector-quantization-based defenses have certifiable robust accuracy and that pRD and swRD demonstrate state-of-the-art performances, surpassing RandDisc by a large margin. Notably, the proposed defenses possess the ``obfuscated gradients" property \cite{Obfuscated}. Our experiments however show that pRD and swRD remain effective under the STE and EOT attacks, which are designed specifically for defenses with gradient obfuscation. These observations confirm that the power of our defenses lies in the effective exploitation of image structure via vector quantization.

\end{abstract}

\end{frontmatter}

\section{Introduction}
\label{intro}
Prevalent as they are, modern deep neural networks (DNNs) are highly vulnerable to adversarial examples \cite{lbfgs}, where carefully crafted small perturbations to the input can cause a well-trained neural network to make incorrect decisions. Since their discovery, adversarial attacks have attracted significant research attention \cite{meng2017, moosavi2017}, leading to intense efforts to develop mechanisms  that protect DNNs from such attacks (see, e.g., \cite{pgd, zhang2021, trades}). 

In image classification tasks, numerous adversarial defense mechanisms have been proposed, including adversarial training \cite{fgsm,pgd,sankaranarayanan2018}, modifications to model architectures \cite{gu2014towards, distillation, zhang2021}, and preprocessing of input images \cite{xu2017feature, randdisc}. Among these different categories of techniques, preprocessing-based approaches are attractive as they do not require re-training the DNN under protection and can be easily integrated to networks that have been deployed. A variety of preprocessing-based adversarial defenses have been introduced and great successes have been demonstrated. 
For example, to preprocess perturbed inputs, Guo et al. \cite{Guo-Obfuscated-preprocessor} apply a range of image transformations such as bit-depth reduction and JPEG compression, while Hu et al. \cite{qiu2021efficient} utilize Discrete Cosine Transform along with image distortion. Among preprocessing-based defenses, Randomized Discretization (``RandDisc",  \cite{randdisc})  is an appealing preprocessing-based defense, due to its simplicity and certifiable robustness. Briefly, RandDisc clusters the pixel values into a set of clusters and replaces each pixel by the value of its corresponding cluster center. 

Despite their successes, preprocessing-based defenses are limited in their resulting adversarial robustness, while often suffering from the ``obfuscated gradients" problem as suggested in \cite{Obfuscated}.  In this work, we aim at developing a novel framework for preprocessing-based adversarial defenses that overcome such limitations. The motivation of this work stems from the recognition that RandDisc can be viewed as image preprocessing through quantization \cite{gersho1992vector}. The key benefit of quantization in this context is that it maps both the original image and its adversarially perturbed version to similar representations, which are subsequently treated almost identically by the downstream classifier. Notably, a quantizer works by partitioning the input space into regions and associating each region with a reproduction point; then, when an input falls within a specific region, it is reconstructed as the reproduction point for that region \cite{cover1999elements}. The optimal quantizers, given the size of the partition \textemdash or equivalently, the number of reconstruction points \textemdash are those that minimize the expected distortion in this reconstruction. Alternatively put, such quantizers partition the entire input space in the most efficient manner. In the context of adversarial defenses, this efficiency translates to effective coverage of all adversarial examples by the partition. 

However, the quantizer used in RandDisc is far from optimal, since it operates on the input image pixel by pixel, that is, performing a form of ``scalar quantization", whereas rate-distortion theory \cite{shannon1959coding} suggests that an optimal quantizer must exploit all input dimensions (i.e., all pixels of the input image) simultaneously so as to maximally exploit the structure of the input distribution. Such a process is referred to as ``vector quantization".

This understanding motivates us to propose a {\em vector quantization framework} for preprocessing-based adversarial defenses. In this framework, the quantizers operate on the images at the ``patch" level rather than at the pixel level as in RandDisc. Specifically, the vector quantizer treats the input image as a collection of patches and, based on the distribution of the patches, finds a set of “representative patches”; each original patch is then modified according to the representative patches close to it. As suggested by rate-distortion theory, this approach promises better exploitation of the inherent structure in images.   In this framework, we develop two new defenses, which we refer to as {\em patched RandDisc (pRD)} and {\em sliding-window RandDisc (swRD)}, where pRD is a direct extension of RandDisc, principled by rate-distortion theory, and swRD is an application-level extension of pRD, designed to enhance practical usability and robustness in real-world scenarios.

\begin{figure}[t]
    \centering
    \includegraphics[width=0.8\linewidth]{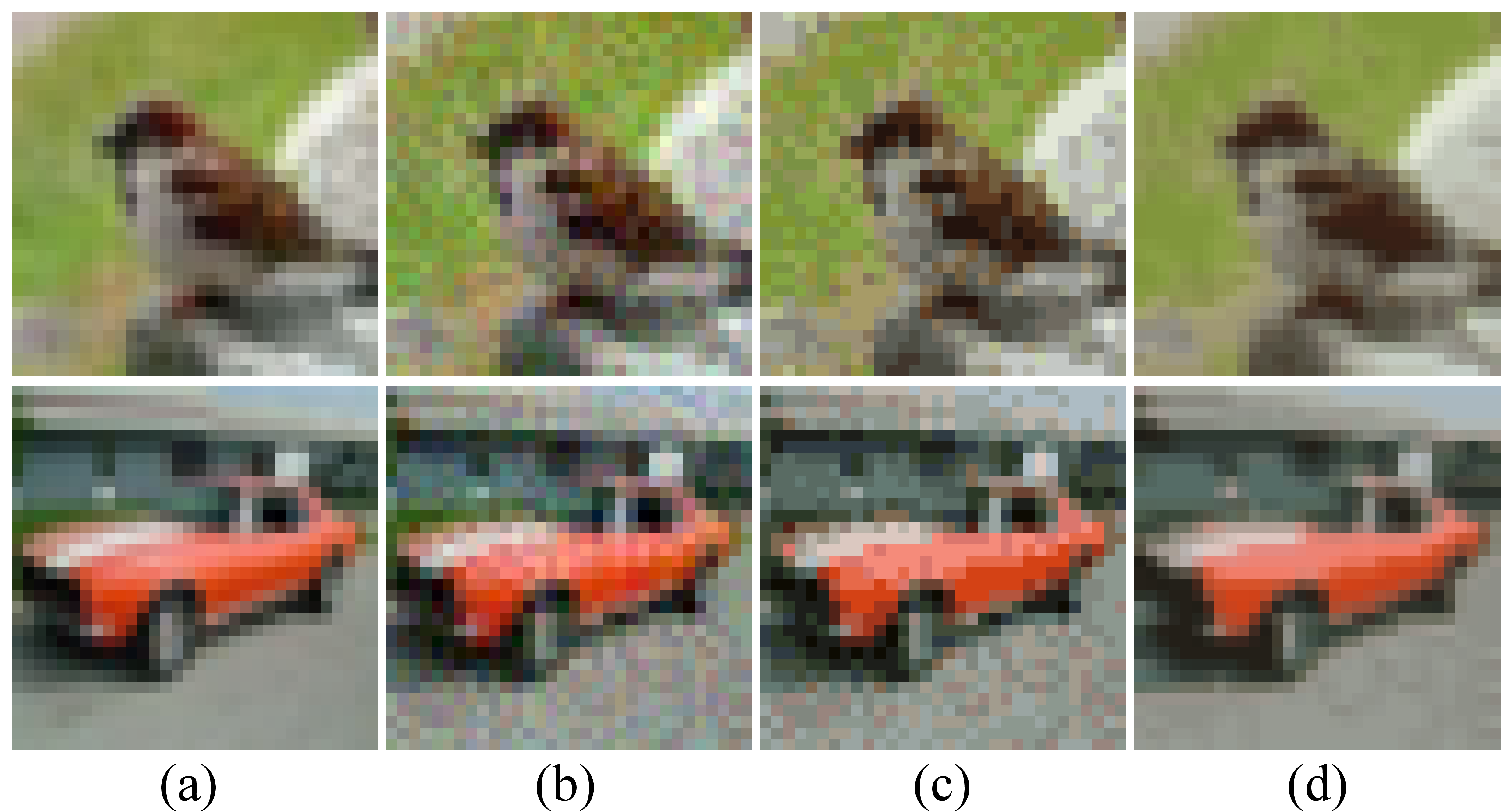}
    \caption{(a) Original images from the CIFAR-10 dataset; (b) adversarially perturbed versions of (a) generated by PGD (perturbation size $\epsilon=16/255$ and $40$ iterative steps); (c) quantized versions of (b) processed by RandDisc;  (d) quantized versions of (b) processed by swRD.}
\label{fig:intro}
\end{figure}

In this work, we show that both pRD and swRD, like RandDisc, offer theoretical certificates of robustness. Through extensive experiments on MNIST, CIFAR-10, CIFAR-100 and ImageNet, we demonstrate that pRD and swRD achieve state-of-the-art robust accuracy against strong attacks (e.g., PGD \cite{pgd}, AutoAttack \cite{autoattack}, and their variants employing STE and EOT \cite{Obfuscated}). Furthermore, we show that pRD and swRD exhibit a higher level of certified robustness compared to RandDisc. Compared with existing adversarial defenses, the proposed defenses simultaneously offer the following additional advantages. 
(1) They are universally applicable to protecting any DNN and are effective for any data distribution. This is because they operate directly on a single image, requiring neither knowledge of the threatened DNN nor access to any dataset.
(2) They are naturally resistant to gradient-based attacks designed specifically for the threatened network protected by these defenses. This is because for any network shielded with these defenses (with the overall network shown in the red dashed box in Figure \ref{fig:diagram}), the gradient signal of the output with respect to the input is zero almost everywhere\footnote{This property is inherent in all quantization-based defenses, including RandDisc.}. 
(3) Arguably the above property is a form of ``obfuscated gradients" as described in \cite{Obfuscated}, where the authors showed that defenses with such a property may not be truly robust against attacks specifically designed for defenses that obfuscate gradient signals. Our experiments however show that even under such attacks, our methods still maintain satisfactory robustness.
(4) Such quantization-based defenses, including our proposed methods as well as RandDisc, may function as an ``image denoiser".  Figure \ref{fig:intro} presents a visual example, where under strong adversarial attacks, our swRD restores images with higher quality than RandDisc does. 

The contributions of this paper are summarized as follows. 
\begin{enumerate}
\item Inspired by RandDisc, we present a vector quantization framework for preprocessing-based adversarial defenses. This framework, theoretically grounded in rate-distortion theory, is universal, free of training, and naturally resistant to gradient-based attacks.
\item We propose two vector quantizers, pRD and swRD, as example defenses in this framework.
\item We show that high-probability certificates of robust accuracy can be obtained for these defenses.
\item We validate these defenses experimentally and demonstrate their state-of-the-art performance. In particular, we show that these defenses do not rely on gradient obfuscation and can resist attacks specifically designed for models and defenses with gradient obfuscation.
\end{enumerate}

The remainder of this paper is organized as follows. Section 2 provides a brief overview of related work on adversarial attacks and defenses. Section 3 presents a detailed review of fundamental results in quantization. Section 4 presents the vector quantization framework and the pRD and swRD defenses therein. In Section 5, we theoretically analyze the certificated robustness of the proposed methods in a high-probability manner. Section 6 reports the experimental results of our approaches in scenarios where the attacks are unaware of the defenses. And Section 7 presents the experimental results where the attacks target the entire network, including the defenses. These results show that the robustness of our defenses does not rely on  gradient obfuscation. Finally, Section 8 summarizes the key contributions of this paper and discusses its limitations and potential improvement.

Throughout this paper, we use capital letters (e.g., $X$ and $Y$) to denote random variables, and lower-case letters (e.g., $x$ and $y$) to denote their specific values.

\section{Related Works}

\textbf{Adversarial Attacks} Adversarial attacks can be categorized into two types: white-box attacks and black-box attacks. In white-box attacks, the adversary has full access to the structure and parameters of the targeted model. In contrast, such information is unavailable in black-box attacks. Projected gradient descent (PGD) \cite{pgd} is one of the most well-known white-box attacks. Given a predefined number of steps and a step size, PGD iteratively updates the input image in the gradient-ascent direction to maximize the loss. Croce and Hein \cite{autoattack} extend PGD to APGD, which eliminates the need of the step size hyperparameter. They further enhance APGD by applying two different loss functions: cross-entropy (CE) and difference of logits ratio (DLR), resulting in two variants of PGD: APGD-CE and APGD-DLR. To strengthen adversarial performance, they combine these two PGD variants with two additional attacks: FAB \cite{FAB} and Square Attack \cite{squareattack}. This ensemble, known as AutoAttack, is parameter-free and independent of manual tuning, demonstrating powerful adversarial capabilities across various experiments.

\textbf{Adversarial Defenses} Adversarial defenses aim to enhance the robustness of DNNs against adversarial attacks. One of the most widely used strategies is adversarial training \cite{fast, trades, mart}, which involves training DNNs on a mix of natural and adversarial examples. Additionally, there are methods focusing on universal robustness against unknown attacks, including meta defenses based on transfer learning \cite{meta-defense}, test-time purification \cite{test-purification}, image noise removal \cite{image-noise-removal}, and random projection filters \cite{rpf}. In addition to these heuristic methods, several defenses provide theoretical certificates \cite{RV5_R1,RV5_R2,RV5_R3,RV5_R4,RV5_R5}, which ensure that a model is either unaffected by adversarial examples within a specified distance from the original input or achieves a maximum robust accuracy on adversarial examples. Randomized smoothing \cite{lecuyer2019certified, cohen2019certified, li2019certified} represents the typical defense offering certified robustness against $\ell_2$ perturbations. Given an arbitrary input, randomized smoothing outputs the label most likely to be predicted by a base classifier under Gaussian noise perturbations of the input. However, recent studies \cite{yang2020randomized, blum2020random} have shown that it fails to achieve nontrivial certified robustness against $\ell_\infty$ perturbations. To address challenges posed by $\ell_\infty$ attacks, Zhang et al. \cite{zhang2021, zhang2021boosting} introduced a novel network architecture incorporating $\ell_\infty$-dist neurons. They proved that the $\ell_\infty$-dist neuron is inherently a 1-Lipschitz function with respect to the $\ell_\infty$ norm. RandDisc is another certified defenses that offers theoretical robustness.

\section{Fundamentals of Quantization}
Quantization is a technique that performs lossy compression of a source, where the source refers to a stochastic process, $X_1, X_2, \dots$. We now briefly review the fundamental theory of quantization, known as rate-distortion theory, for i.i.d. sources,  and then discuss its implications in image compression.

Let ${\cal X}$ denote the space from which each $X_i$ takes its value. A length-$n$ quantizer consists of a pair of functions $(f_n, g_n)$; $f_n$, called the {\em encoder}, maps a block of $n$ symbols $(X_1, X_2, \ldots, X_n)$ to an integer in $\{1, 2, \ldots, M\}:=[M]$, and $g_n$, called the {\em decoder}, maps an integer in $[M]$ (the output of the encoder) to a reconstructed sequence $(\hat{X}_1, \hat{X}_2, \ldots, \hat{X}_n$), where each $\hat{X}_i$ lives in the space $\hat{\cal X}$ (usually a subset of ${\cal X}$). The {\em rate} of the quantizer $(f_n, g_n)$ is defined as $\frac{\log M}{n}$, indicating the average number of bits used to compress each source symbol (when the logarithm is taken as base 2).

There is a non-negative function $d$ on ${\cal X}\times \hat{\cal X}$, called the distortion function, that measures some notion of ``distortion" or ``error" between a symbol $x\in {\cal X}$ and its reconstructed version $\hat{x}\in \hat{\cal X}$. The sequence-level distortion function $d_n$ induced by $d$ is defined by 
\[
d_n((x_1, x_2, \ldots, x_n),(\hat{x}_1, \hat{x}_2, \ldots, \hat{x}_n)):= \frac{1}{n}\sum\limits_{i=1}^n d(x_i, \hat{x}_i).
\]

A fundamental concept in quantization is the rate-distortion function $R(D)$, defined as the infimum of all rates achievable by any quantizer $(f_n, g_n)$ (across all $n$'s) with expected sequence-level distortion no greater than $D$. In his pioneering work \cite{shannon1959coding}, Shannon proved that 
\[
R(D) = \min_{P_{\hat{X}^|X}: E d(X, \hat{X})\le D} I(X; \hat{X}).
\]
where $I(X; \hat{X})$ is the mutual information between $X$ and $\hat{X}$ induced by the source distribution $P_X$ and a conditional distribution $P_{\hat{X}|X}$ (which gets minimized over in the equation above). 

In the context of this paper, it is convenient to regard the quantizer $(f_n, g_n)$ as the composition $g_n\circ f_n$,  mapping a source sequence $(x_1, \ldots, x_n)$ directly to its reconstruction (or reproduction) $(\hat{x}_1, \ldots, \hat{x}_n)$. In this view, the quantizer $g_n\circ f_n$ operates on the space ${\cal X}^n$ and induces a partition of ${\cal X}^n$ into a set of $M$ disjoint regions $\{{\cal R}_1, {\cal R}_2, \ldots {\cal R}_M\}$. Each region ${\cal R}_i$ is associated with a reconstruction point $g_i \in {\cal R}_i$, and we refer to the set $\{({\cal R}_i, g(i)):i\in [M]\}$ of region-reconstruction pairs as the quantization codebook. When an input  $(x_1, x_2, \ldots, x_n)$ falls in ${\cal R}_i$, it is reconstructed as the point $g(i) \in \hat{\cal X}^n$. When $n=1$, $g_n\circ f_n$ compresses the source symbol-by-symbol and is called a scalar quantizer. When $n>1$, $g_n\circ f_n$ compresses the source  block-by-block, where each block contains $n$ symbols. Such a quantizer is referred to as a vector quantizer. 


In rate-distortion theory, even for i.i.d. sources, it is well known that achieving  the fundamental limit given by the rate-distortion function generally requires the use of a vector quantizer with length $n$ going to infinity. Thus, scalar quantizers are usually far from optimal, and to approach the optimal performance, one must use a quantizer with a large length. For non-i.i.d. sources, where dependencies exist across symbols, vector quantizers are evidently even more favorable, since the dependency naturally implies the necessity of jointly compressing multiple symbols at once. 

In lossy image compression, we may view an image with $m$ pixels as a realization of a random vector $(X_1, X_2, \ldots, X_m)$, where each $X_i$ corresponds to a pixel and there are dependencies across pixels. When the distribution of this random vector is known, the optimal compression is only achievable with a vector quantizer with length $n=m$, and the optimal quantizers must exploit the distribution of the random vector. When we are given only a single image without access to the distribution of $(X_1, X_2, \ldots, X_m)$, a more practical approach is to group the $m$ pixels into patches, and approximate the patches as i.i.d.. Then when there are adequate number of patches, we may have good approximation of the distribution of a patch and perform quantization patch by patch. The quantizers proposed in this paper essentially adopt this strategy. 

\section{Defenses with Vector Quantization}
\label{sec:3}

\begin{figure}[t]
  \centering
  \includegraphics[width=0.8\linewidth]{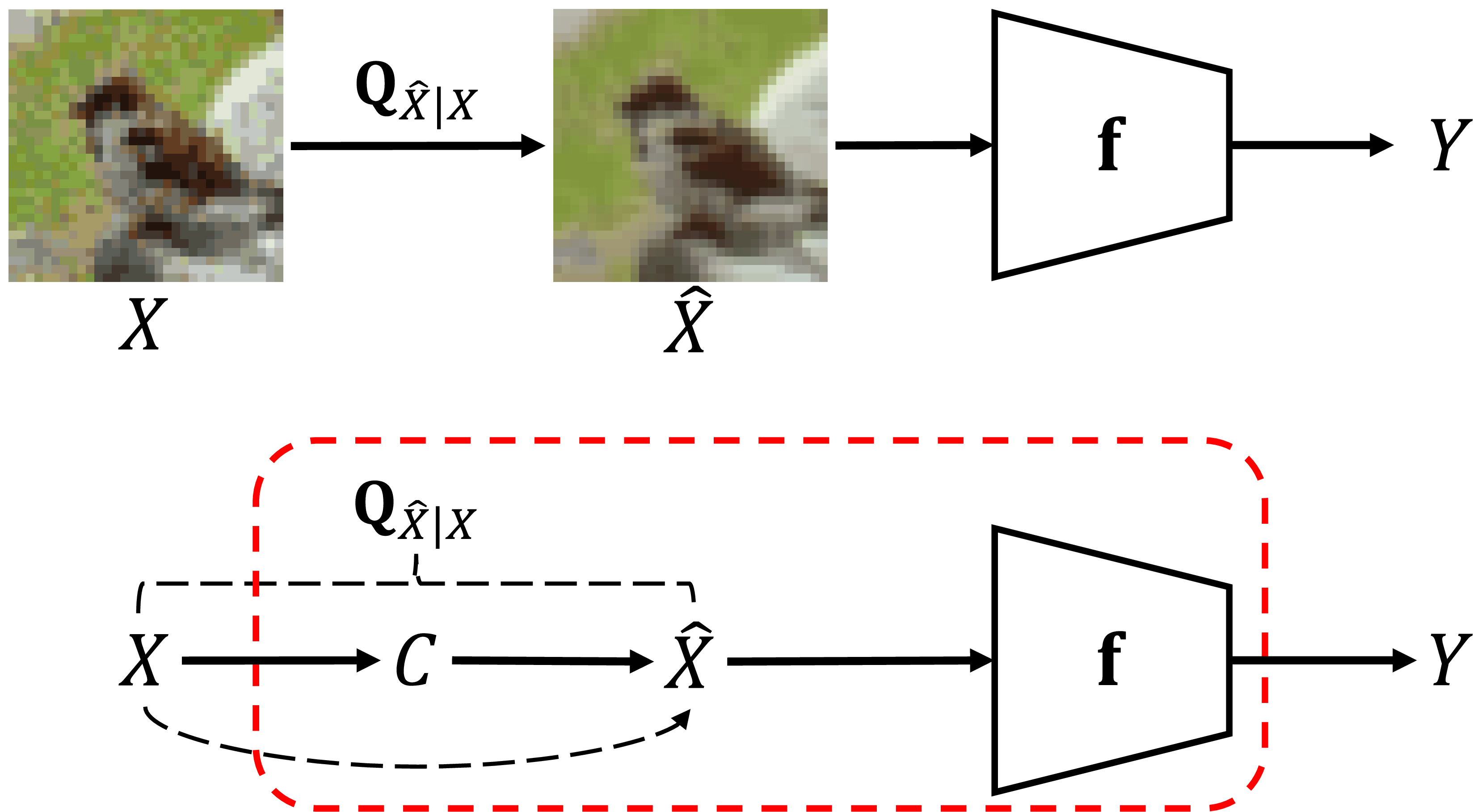}
  \caption{Top: the overall framework of pre-processing based  adversarial defenses, where $X$ is the input, $\mathbf{Q}_{\hat{X}|X}$ is the (possibly randomized) preprocessor, and  $\hat{X}$ is the preprocessed input, ${\bf f}$ is the classifier to be protected, $Y$ is the predicted label. Bottom: Quantization-based defenses belong to the preprocessing-based defenses, in which the preprocessor $\mathbf{Q}_{\hat{X}|X}$ is implemented as a quantizer with $C$ being its set of reproduction points. The proposed quantizers pRD or swRD both have such a structure.}
\label{fig:diagram}
\end{figure}

We now present the vector quantization framework as preprocessing-based adversarial defenses. In this framework, an input image $X$ is replaced with its quantized counterpart $\hat{X}$ through a vector-quantizer $\mathbf{Q}$ specified by a conditional distribution $\mathbf{Q}_{\hat{X}|X}$. At a high level, the vector quantizer $\mathbf{Q}$ first treats the input image as a collection of patches (possibly overlapping), and based on the distribution of the patches, finds a set of ``representative patches", or reproduction points. Each original patch is then modified according to the patches close to it. The quantized image is then obtained by assembling these modified patches.
Such a quantizer can then be used to preprocess the input image before it is passed to the classifier threatened by adversarial attacks.  The overall architecture of this framework is also explained in Figure \ref{fig:diagram}. 

We introduce two example defenses in this framework: patched RandDisc (pRD) and sliding-window RandDisc (swRD), with pseudocode given in Algorithms \ref{alg-prd} and \ref{alg-swrd}.

\begin{figure}[t]
    \centering
    \includegraphics[width=0.8\linewidth]{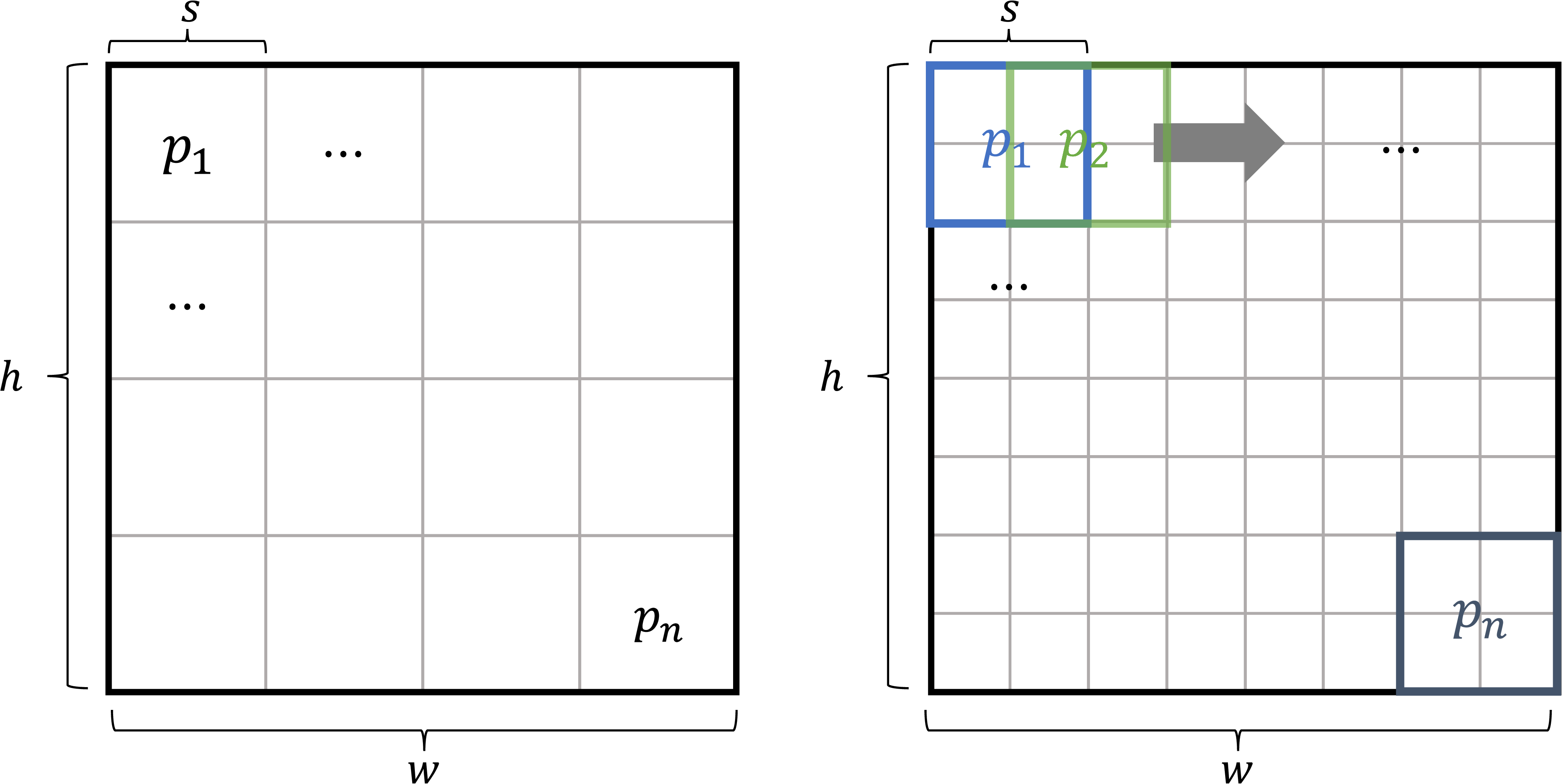}
    \caption{Patch generation in pRD (left) and swRD (right), where $h$ and $w$ denote image size, and $s$ indicates patch size in pRD and window size in swRD.}
\label{fig:partation}
\end{figure}
\subsection{Quantizer $\mathbf{Q}_{\hat{X}|X}^{\rm pRD}$ in pRD}
\label{sec:pRD}
Given an input image $x$ of size $h\times w$, the quantizer $\mathbf{Q}_{\hat{X}|X=x}^{\rm pRD}$ is defined as follows:

{\bf 1. Finding reproduction points.} The image $x$ is first divided into a set $\mathcal{P}:=\{p_1,\dots,p_n\}$ of $n$ non-overlapping patches, all of equal size, as illustrated in Figure \ref{fig:partation} (left). If each patch contains $s\times s$ pixels and $q$ channels, then each patch $p_i$ is represented as a vector in $\mathbb{R}^{qs^2}$, and the total number of patches in $\mathcal{P}$ is given by $n=(h/s)(w/s)$. If necessary, the image is zero-padded to ensure $h/s$ and $w/s$ are integers. 
Next, independent Gaussian noise $\epsilon_i\sim\mathcal{N}(0, \sigma^2 \mathbf{I})$ is added to patch $p_i$ for $i=1,\dots,n$, where $\mathbf{I}$ is a $qs^2\times qs^2$ identity matrix. These noisy patches are then clustered into $k$ clusters, giving rise to a set $\mathcal{C}:=\{c_1,\dots, c_k\}$ of cluster centers. These cluster centers serve as the ``reproduction points" of the quantization codebook. 
Any clustering algorithm can be applied for this step. While RandDisc employs k-means++ \cite{kpp}, we simply choose k-means algorithm in this work. 

{\bf 2. Image reconstruction using reproduction points.} The quantized version $\hat{p}_i$ of each patch $p_i$ is determined by finding the closest cluster center to $p_i+\omega_i$, where $\omega_i\sim\mathcal{N}(0, \tau^2 \mathbf{I})$. Mathematically, this is expressed as:
\begin{equation}
\label{equ:replace_prd}
    \hat{p}_i = \argmin_{c\in \mathcal{C}} \Vert p_i+\omega_i-c\Vert_2.
\end{equation}

The quantized patches $\hat{p}_1,\dots,\hat{p}_n$ are used to replace their corresponding original patches $p_1,\dots,p_n$, resulting in the quantized image $\hat{x}$. 

\begin{center}  
\begin{minipage}{.9\linewidth}  
\begin{algorithm}[H]
    \caption{pRD}
    \label{alg-prd}
    \begin{algorithmic}[1]
        \REQUIRE An image $x$ of size $h\times w$, the patch size $s$, the number $k$ of clusters, and the noise levels $\sigma$, $\tau$.
        \ENSURE The quantized version $\hat{x}$ of $x$.
        \STATE Initialization: $n=(h/s)(w/s)$, $\mathcal{C}=\emptyset$, and $\omega_i\overset{\rm iid}{\sim} \mathcal{N}(0, \tau^2 {\bf I})$, $ \epsilon_i \overset{\rm iid}{\sim}\mathcal{N}(0, \sigma^2 {\bf I})$ for $i=1,\dots,n$.
        \STATE $\{p_1,\dots,p_n\} \leftarrow$ partition $x$ into $n$ clusters without overlap 
        \STATE $\mathcal{C} \leftarrow$ cluster the set $\{p_1 +\epsilon_1,\dots,p_n +\epsilon_n\}$ into $k$ cluster centers
        \FOR{$i=1$ to $n$}
            \STATE $\hat{p}_i\leftarrow \argmin_{c\in \mathcal{C}} \Vert p_i+\omega_i-c\Vert_2$ \COMMENT{the closest cluster center}
        \ENDFOR
        \STATE $\hat{x}\leftarrow $ reconstruct $n$ quantized patches $\{\hat{p}_1,\dots,\hat{p}_n\}$ into an image                
        \ENSURE  $\hat{x}$
    \end{algorithmic}  
\end{algorithm}
\end{minipage}
\end{center}

\subsection{Quantizer $\mathbf{Q}_{\hat{X}|X}^{\rm swRD}$ in swRD}
\label{sec:swRD}
{\bf 1. Finding reproduction points.} For an input image $x$ of size $h\times w$, the quantizer $\mathbf{Q}_{\hat{X}|X=x}^{\rm swRD}$ first slides a window of size $s\times s$ over $x$ with a stride of one pixel. At each position, the pixels within the window form a patch $p_i$. This process generates a set $\mathcal{P}$ of $n$ overlapping patches, denoted as $\mathcal{P}:=\{p_1,\dots,p_n\}$. The total number $n$ of patches in $\mathcal{P}$ is given by $n=(h-s+1)(w-s+1)$. A visual representation of this process is shown in Figure \ref{fig:partation} (right). As in pRD, a set $\mathcal{C}$ of $k$ cluster centers (reproduction points) is subsequently obtained by k-means algorithm. 

{\bf 2. Image reconstruction using reproduction points.} The quantized version $\hat{p}_i$ of each patch $p_i$ is computed as the closest cluster center to the noisy patch $p_i+\omega_i$. Due to the overlapping nature of patches, the reconstruction process differs from the patch-by-patch replacement used in pRD. Instead, a pixel-by-pixel approach is employed, leveraging the fact that a pixel is covered in multiple patches. The process of pixel replacement is described in detail below:
Let $x_i$ be the $i$-th pixel in the image $x$. Typically, $x_i$ is covered by $s^2$ patches, except for pixels near the edges or corners of the image. Define $\mathcal{P}^i:=\{\hat{p}^i_1,\hat{p}^i_2,\dots\}$ as the set of quantized patches that cover pixel $x_i$, and $\mathcal{S}^i:=\{s^i_1,s^i_2,\dots\}$ as the set of similarity values between each quantized patch $\hat{p}^i_j$ in $\mathcal{P}^i$ and its corresponding original version $p^i_j$. The similarity values are computed by
\begin{equation}
\label{equ:replace_prd}
    s^i_j= -\beta\Vert\hat{p}^i_j -p^i_j\Vert_2^2.
\end{equation}

This process of obtaining $\mathcal{P}^i$ and $\mathcal{S}^i$ corresponds to lines 4-6 and 9-14 in Algorithm \ref{alg-swrd}. Then the pixel $x_i$ is replaced by a weighted sum of the corresponding pixels in each quantized patch from $\mathcal{P}^i$. Let $\hat{p}^{i*}_j$ denote the pixel in $\hat{p}^i_j$ that corresponds to the position of $x_i$. The new pixel value $\hat{x}_i$ is calculated as:
\begin{equation}
\begin{aligned}
\label{equ:replace_swrd}
    w^i_j &= \frac{\exp{(s^i_j)}}{\sum_{l=1}^{{\rm len}(\mathcal{S}^i)}\exp{(s_l^i)}},\\
    \hat{x}_i &= \sum_{j=1}^{{\rm len}(\mathcal{P}^i)}\hat{p}^{i*}_j w^i_j.
\end{aligned}
\end{equation}

Finally, the new pixels values $\hat{x}_1,\dots,\hat{x}_m$ are used to assemble the quantized image $\hat{x}$.

\begin{center}  
\begin{minipage}{.9\linewidth}  
\begin{algorithm}[H]
    \caption{swRD}
    \label{alg-swrd}
    \begin{algorithmic}[1]
        \REQUIRE An image $x$ with size $h\times w$, the patch size $s$, the number $k$ of clusters, the parameter $\beta$, and the noise levels $\sigma$, $\tau$.
        \ENSURE The quantized version $\hat{x}$ of $x$.
        \STATE Initialization: $n=(h-s+1)(w-s+1)$, $m=hw$, $\mathcal{C}=\mathcal{P}^i=\mathcal{S}^i=\emptyset$ for $i=1,\dots,m$, and $\omega_j\overset{\rm iid}{\sim} \mathcal{N}(0, \tau^2 {\bf I})$, $ \epsilon_j \overset{\rm iid}{\sim}\mathcal{N}(0, \sigma^2 {\bf I})$ for $j=1,\dots,n$.
        \STATE $\{p_1,\dots,p_n\} \leftarrow$ partition $x$ into $n$ clusters with overlap
        \STATE $\mathcal{C} \leftarrow$ cluster the set $\{p_1+\epsilon_1,\dots,p_n+\epsilon_n\}$ into $k$ cluster centers 
        \FOR{$i=1$ to $n$}
            \STATE $\hat{p}_i\leftarrow \argmin_{c\in \mathcal{C}} \Vert p_i+\omega_i-c\Vert_2$ \COMMENT{the closest cluster center}
            \STATE $d_i\leftarrow \Vert\hat{p}_i-p_i\Vert_2$\COMMENT{the distance}
        \ENDFOR
        \FOR{$i=1$ to $m$}
            \FOR{$j=1$ to $n$}
                \IF{patch $\hat{p}_j$ covers pixel $x_i$}
                    \STATE $\mathcal{P}^i\leftarrow \mathcal{P}^i\cup \hat{p}_j$
                    \STATE $\mathcal{S}^i\leftarrow \mathcal{S}^i\cup \{-\beta d_j^2\}$\COMMENT{the similarity set}
                \ENDIF \COMMENT{$\mathcal{P}^i=\{p_1^i,p_2^i,\dots\}$ and $\mathcal{S}^i=\{s_1^i,s_2^i,\dots\}$}
            \ENDFOR
            \STATE $w^i_j \leftarrow \frac{\exp{(s^i_j)}}{\sum_{l=1}^{{\rm len}(\mathcal{S}^i)} \exp{(s^i_l)}}$\COMMENT{the weight of patch $j$ in $\mathcal{P}^i$}
            \STATE $\hat{x}_i \leftarrow \sum_{j=1}^{{\rm len}(\mathcal{P}^i)}p^{i*}_j w^i_j$ \COMMENT{the weighted sum of all patches in $\mathcal{P}^i$}
        \ENDFOR
        \STATE $\hat{x}\leftarrow $ reconstruct $m$ quantized pixels $\{\hat{x}_1,\dots,\hat{x}_m\}$ into an image                
        \ENSURE  $\hat{x}$
    \end{algorithmic}  
\end{algorithm}
\end{minipage}
\end{center}
\subsection{Using $\mathbf{Q}_{\hat{X}|X}$ as Adversarial Defenses}   

Let $x$ denote a natural image and $x'$ its adversarially perturbed image. Let $\hat{x}$ and $\widehat{x'}$ denote their respective quantized versions. Let $c$ and $c'$ respectively represent the set of cluster centers obtained from $x$ and that obtained from $x'$, and let $\mathbf{f}$ represent a classifier that is subject to adversarial attacks. The adversarial attacks considered in this paper are projected gradient descent (PGD) \cite{pgd} and AutoAttack \cite{autoattack}, where PGD is one of the most powerful white-box attacks and AutoAttack is an ensemble of white-box and black-box attacks. These attacks generate an adversarial example $x'$ from a natural image $x$ under $\epsilon$-bounded $\ell_\infty$-norm constraint, i.e., $||x-x'||_\infty\leq \epsilon$.

To protect classifier $\mathbf{f}$ from such attacks, an input image $X$ is first processed by the quantizer $\mathbf{Q}_{\hat{X}|X}$, and then passed to model $\mathbf{f}$, as shown in Figure \ref{fig:diagram}.  For later use, we denote by $\mathbf{Q}_{C|X}$ the probabilistic mapping that generates the set $C$ of cluster centers from the input image $X$. Likewise, $\mathbf{Q}_{\hat{X}|C,X}$ denotes the mapping that converts the image $X$ to $\hat{X}$ based on $C$.

In both pRD and swRD, adding isotropic Gaussian noise makes the  defenses stochastic, so that the output image can be viewed as a random variable following a conditional distribution  $\mathbf{Q}_{\hat{X}|X}$ given an input image $X$. Without such randomization, the algorithm is a deterministic function of the input image, potentially more sensitive to input perturbations. Notably, when the input is an adversarial example $x'$, which could lie anywhere within the $\ell_\infty$ ball centered at $x$ with radius $\epsilon$, the deterministic function may not protect the target model against all adversarial examples in this ball. In contrast, stochastic mappings can provably defend the target model against all adversarial examples with high probability, as we will show subsequently. On the other hand, there appears no practical gain in the second injection of noise as we show in \ref{apdx: adding noise once}. However, it does provide convenience in mathematical analysis and offer a theoretical certificates, as we next discuss.

\section{High-Probability Certification of Robustness}
\label{sec:kld}

In this section, we show that high-probability certificates of robustness can be established for quantization-based  preprocessing defenses. We begin by introducing  several definitions.
\subsection{Definitions}

We consider a general $K$-class classification problem, where the target classifier $\mathbf{f}$ maps each $x$ to a distribution over the $K$ labels. For each class label $y$, we use $\mathbf{f}(x)[y]$ to denote the probability that $\mathbf{f}$ predicts the label of $x$ is $y$. We will use $\mathcal{A}$ to denote an arbitrary adversarial attack.

\begin{defn}[Classification Margin]
For a given instance-label pair $(x, y)$, the classification margin $\rho(x,y,\mathbf{f})$ is 
defined as
\begin{equation}
\label{equ:margin}
\begin{aligned}
    \rho(x,y,\mathbf{f}) := \mathbf{f}(x)[y] - \max_{j\neq y} \mathbf{f}(x)[j].
\end{aligned}
\end{equation}
\end{defn}

In a nutshell, $\rho(x,y,\mathbf{f})$ is the ``relative confidence" that $\mathbf{f}$ labels $x$ with $y$, with respect to the confidence that it labels $x$ with its most confident other label. When $y$ is the true label, this quantity indicates a ``margin" that $f$ makes a mistake for $x$; the prediction for $x$ is correct if and only if this margin is positive. The following definition is then natural.

\begin{defn}[True Robust Error]
Let $\mathbf{Q}_{\hat{X}|X}$ be a randomized defense for a classifier $\mathbf{f}$. That is,  for any input $x$, the protected classifier outputs $\mathbf{f}(\hat{x})$, where $\hat{x}$ is drawn from $\mathbf{Q}_{\hat{X}|X=x}$. For a given instance-label pair $(x, y)$ where $y$ is the true label of $x$, the true robust error of $x$ under attack $\mathcal{A}$ for classifier $\bf{f}$ defended by $\mathbf{Q}_{\hat{X}|X}$  is defined by
\begin{equation}
\begin{aligned}
    {\rm Err}(x,y|\mathcal{A},\mathbf{f},\mathbf{Q}_{\hat{X}|X}) := \Pr_{\widehat{x'}\sim \mathbf{Q}_{\hat{X}|X=\mathcal{A}(x)}}[\rho(\widehat{x'},y,\mathbf{f})\leq0].
\label{eq: generalization-robust-risk}
\end{aligned}
\end{equation}
\end{defn}

\subsection{Certification of Robust Accuracy}
\label{sec: pro1&2}
\begin{prop}
For any input $x$, let $\{\widehat{x'}_i\}_{i=1}^{m}$ be a set of quantized perturbed images drawn i.i.d. from $\mathbf{Q}_{\hat{X}|X=\mathcal{A}(x)}$. If $x$ is correctly labeled by $\mathbf{f}$ as $y$, then with probability at least $1-\delta$, 
\begin{equation}
\begin{aligned}
\label{equ:pro1}
    {\rm Err}(x,y|\mathcal{A},\mathbf{f},\mathbf{Q}_{\hat{X}|X}) \leq \frac{\frac{1}{m}\sum_{i=1}^{m}\Vert\mathbf{f}(x)-\mathbf{f}(\widehat{x'}_{i})\Vert_1 + \sqrt{(2\log{\frac{1}{\delta}})/m}}{\rho(x,y,\mathbf{f})}.
\end{aligned}
\end{equation}
\end{prop}

The proof of this proposition is provided in \ref{apdx: proof P1}. This result allows us to provide high-probability robustness certification for the defense $\mathbf{Q}_{\hat{X}|X}$ by repeatedly sampling $m$ images $\widehat{x'}_1, \widehat{x'}_2, \ldots, \widehat{x'}_m$ from $\mathbf{Q}_{\hat{X}|X=\mathcal{A}(x)}$ and measuring $\frac{1}{m}\sum_{i=1}^{m}\Vert\mathbf{f}(x)-\mathbf{f}(\widehat{x'}_{i})\Vert_1$ and $\rho(x, y, \mathbf{f})$.  
For example,  suppose $m=100$, $\frac{1}{m}\sum_{i=1}^{m}\Vert\mathbf{f}(x)-\mathbf{f}(\widehat{x'}_{i})\Vert_1=0.2$, and $\rho(x,y,\mathbf{f})=0.9$. Then with probability 0.95 (i.e., $\delta=0.05$), the true robust error
is certified to be no greater than $0.40$, namely, the 95\%-certified robust accuracy is $0.60$; with probability 0.995 (i.e., $\delta=0.005$) the true robust error is certified to be no greater than $0.45$ and the 99.5\%-certified robust accuracy is $0.55$.

In \cite{randdisc}, the authors show that when $\mathbf{f}$ correctly classifies $x$, the KL divergence ${\rm KL}(\mathbf{Q}_{\hat{X}|X=x}||\mathbf{Q}_{\hat{X}|X=\mathcal{A}(x)})$ can also be used to certify the randomized defense $\mathbf{Q}_{\hat{X}|X}$ for the input $x$. Specifically, smaller KL divergence leads to a higher confidence level for a certification with the 100\% robust accuracy.  Indeed, if the distributions $\mathbf{Q}_{\hat{X}|X=x}$ and $\mathbf{Q}_{\hat{X}|X=\mathcal{A}(x)}$ are close, then when the quantized version of $x$ can be correctly classified by $\mathbf{f}$, so will the quantized version of $\mathcal{A}(x)$.

As the exact KL divergence ${\rm KL}(\mathbf{Q}_{\hat{X}|X=x}||\mathbf{Q}_{\hat{X}|X=\mathcal{A}(x)})$
can be difficult to compute, the authors of  \cite{randdisc} use an upper bound on the KL divergence instead. Inspired by \cite{randdisc}, in a later section, we will also use KL divergence as an indirect measure of the certified robust accuracy. In particular, we will use a tighter upper bound on the KL divergence ${\rm KL}(\mathbf{Q}_{\hat{X}|X=x}||\mathbf{Q}_{\hat{X}|X=\mathcal{A}(x)})$,  defined by 
\begin{equation}
\begin{aligned}
\label{eq:u-kl}
    U_{\rm KL}(x): = {\rm KL}(\mathbf{Q}_{C|X=x}&||\mathbf{Q}_{C|X=\mathcal{A}(x)})  \\
    &+\mathbb{E}_{c\sim \mathbf{Q}_{C|X=x}}[{\rm KL}(\mathbf{Q}_{\hat{X}|C=c,X=x}||\mathbf{Q}_{\hat{X}|C=c,X=\mathcal{A}(x)})].
\end{aligned}
\end{equation}

Due to the data-processing inequality \cite{cover1999elements},  we have
\begin{equation}
\label{equ:kl-c}
    {\rm KL}(\mathbf{Q}_{\hat{X}|X=x}||
    \mathbf{Q}_{\hat{X}|X=\mathcal{A}(x)}) \leq U_{\rm KL}(x).
\end{equation}

Note that our upper-bound $U_{\rm KL}(x)$ is tighter than that used in \cite{randdisc} as the latter uses the data-processing inequality twice.

In this work, we develop a numerical approach to estimate this upper bound $U_{\rm KL}(x)$, given in Section \ref{sec: exp-p1}, and we show, in Section \ref{sec:exp-kld},  that pRD and swRD exhibit much smaller values of this bound. Notably, the improvement of pRD and swRD over RandDisc in this bound primarily lies in the improvement in the first term of the bound, namely, ${\rm KL}(\mathbf{Q}_{C|X=x}||\mathbf{Q}_{C|X=\mathcal{A}(x)})$. This improvement can be intuitively explained as follows: compared to RandDisc, pRD and swRD perform vector quantization in a higher dimensional space. In this space, there is arguably more structure in the signal distribution. The cluster centers $c$ obtained from $x$ are more resilient to noise or perturbation. Details of the estimation approach and experimental results are presented in Sections \ref{sec: exp-p1} and \ref{sec:exp-kld}.

\section{Experiments: Attacks Ignorant of Defenses}
\label{sec 4}

In this section, we conduct experiments in a scenario where the adversaries are unaware of the existence of defenses. Specifically, the adversaries compute the gradient at the input $x_0$ using $\triangledown_x {\bf f} (x)\big|_{x=x_0}$, and perturb the input in the direction of the gradient. In the next section, we will consider a different scenario where the adversaries attack the composition of the defenses and the classifier.

All experiments are implemented using PyTorch \cite{pytorch}. We evaluate the performance of pRD and swRD against two adversarial attacks: PGD \cite{pgd} and AutoAttack \cite{autoattack}.  Each reported accuracy value for our methods is the average from 10 independent runs, presented together with the standard deviation (in parentheses within the tables).

The results for our defenses and  RandDisc are based on their respective optimal settings of $(s, k)$. Additionally, we experimentally compare the certified robustness of RandDisc, pRD, and swRD.
\subsection{Evaluation of Robust Accuracy}
\label{sec:exp-robust acc}
\begin{table}[H]
    \centering
    \caption{Comparison of accuracy (\%) between our methods and prior defenses on the MNIST dataset with perturbation size $\epsilon=0.31$.}
    \label{tab:mnist}
    \begin{center}
    
        \begin{tabular}{l|c|cc}
        \hline
          \multirow{2}*{\textbf{Defense}}  & \multirow{2}*{$\mathbf{Acc_{nat}}$} & \multicolumn{2}{c}{$\mathbf{Acc_{rob}}$}\\
          
           ~    & ~ & PGD  & AutoAttack   \\
        \hline\hline
        Trades             &\textbf{99.38}     &95.07     &91.61     \\
        
        $A^5/O$            &98.65    &96.94     &\textbf{95.88}             \\
        
        RandDisc
        ($s$=1, $k$=2)           
        &99.23(0.02)  &75.97(0.15)  &92.36(0.12)              \\
        \hline
        pRD
        ($s$=2, $k$=2)    
        &97.75(0.04)  &92.29(0.14)  &85.84(0.13)            \\
        
        swRD
        ($s$=2, $k$=2)  
        &99.26(0.01)  &\textbf{97.07}(0.06)  &91.96(0.07)    \\
        \hline
        \end{tabular}
    \end{center}
    
\end{table}

\begin{table}[H]
    \centering
    \caption{Comparison of accuracy (\%) between our methods and prior defenses on the CIFAR-10 dataset with perturbation size $\epsilon=8/255$. }
    \label{tab:cifar10}
    \begin{center}
    
    \begin{tabular}{l|c|cc} 
    \hline
      \multirow{2}*{\textbf{Defense}}   &  $\mathbf{Acc_{nat}}$ & \multicolumn{2}{c}{$\mathbf{Acc_{rob}}$}\\
      
       ~      & ~ & PGD  & AutoAttack   \\
    \hline\hline
    Trades    &83.00  &53.18  &49.21     \\
    PORT      &86.24  &63.70  &60.21    \\
    BS        &82.38  &51.67  &49.78    \\
    SCORE     &86.79  &63.47  &59.87    \\
    AdaAD     &86.75  &54.13  &50.06    \\
    $A^5/O$   &51.91  &50.98  &49.14             \\
    WANG      &\textbf{90.27}    &67.28  &65.45    \\
    RandDisc
    ($s$=1, $k$=25)  &89.58(0.13)  &10.82(0.29)  &13.92(0.41)               \\
    \hline
    pRD
    ($s$=2, $k$=25)  &79.43(0.11)  &63.93(0.31)  &60.43(0.44)          \\
    pRD
    ($s$=2, $k$=70)  &88.18(0.13)  &36.46(0.54)  &32.24(0.63)          \\
    swRD
    ($s$=3, $k$=25)  &83.42(0.13)  &\textbf{75.56}(0.21)  &\textbf{73.67}(0.38)    \\
    swRD
    ($s$=3, $k$=70)  &88.37(0.12)  &70.44(0.21)  &67.64(0.43) 
    \\
    \hline
    \end{tabular}
    \end{center}
\end{table}

\begin{table}[H]
    \centering
    \caption{Comparison of accuracy (\%) between our methods and prior defenses on the CIFAR-100 dataset with perturbation size $\epsilon=8/255$. }
    \label{tab:cifar100}
    \begin{center}
    \begin{tabular}{l|c|cc}
    \hline
      \multirow{2}*{\textbf{Defense}}  &  $\mathbf{Acc_{nat}}$ & \multicolumn{2}{c}{$\mathbf{Acc_{rob}}$}\\
      
       ~   & ~ & PGD  & AutoAttack   \\
    \hline\hline
    Trades      &57.82  &30.38  &25.02     \\
    PORT        &64.97  &35.60  &30.87    \\
    BS          &57.58  &30.67  &26.24    \\
    SCORE       &57.58  &30.67  &26.24   \\
    AdaAD       &62.19  &32.52  &26.74    \\
    WANG        &\textbf{78.60}  &40.80  &38.80    \\
    RandDisc
    ($s$=1, $k$=25) &70.06(0.38)  &8.55(0.21)  &16.52(0.47)        \\
    \hline
    pRD
    ($s$=2, $k$=25) &57.71(0.17)  &40.81(0.30)  &43.64(0.44)         \\
    pRD
    ($s$=2, $k$=90) &70.87(0.16)  &12.03(0.17)  &19.36(0.38)         \\
    swRD
    ($s$=3, $k$=35) &60.63(0.33) &\textbf{50.67}(0.31) &\textbf{53.78}(0.46)  \\
    swRD
    ($s$=3, $k$=90) &70.04(0.19) &42.02(0.24) &50.09(0.47)

    \\
    \hline
    \end{tabular}
    \end{center}
\end{table}

\begin{table}[H]
   
    \centering
    \caption{Comparison of accuracy (\%) between our methods and prior defenses on the ImageNet dataset with perturbation size $\epsilon=4/255$. }
    \label{tab:imagenet}
    \begin{center}
        \begin{tabular}{l|c|c}
        \hline
      \multirow{2}*{\textbf{Defense}}   &  $\mathbf{Acc_{nat}}$ & $\mathbf{Acc_{rob}}$\\
      
       ~           & ~ & AutoAttack   \\
    \hline\hline
    STL            &68.30  &50.20      \\
    Salman         &64.00  &35.00    \\
    RandDisc
    ($s$=1, $k$=30) &67.69(1.02)  &46.49(0.52)                 \\
    \hline
    pRD
    ($s$=2, $k$=30) &65.56(0.98)  &62.36(0.97)          \\
    swRD
    ($s$=3, $k$=30) &\textbf{69.32}(0.33)  &\textbf{67.50}(0.99)      \\
    \hline
    \end{tabular}
    \end{center}
\end{table}

In our experiments, PGD and AutoAttack are selected as adversarial attacks since they represent state-of-the-art attack schemes. The datasets used include MNIST \cite{mnist}, CIFAR-10, CIFAR-100 \cite{cifar} and ImageNet \cite{imagenet}. We evaluate both natural accuracy ($\text{Acc}_{\rm nat}$, where $\epsilon=0$) and robust accuracy ($\text{Acc}_{\rm rob}$, where $\epsilon>0$) of pRD and swRD on the test datasets. These results are compared with several current state-of-the-art adversarial defense techniques, including Trades \cite{trades}, $A^5/O$ \cite{A5o}, PORT \cite{port}, Backward Smoothing \cite{back_smoothing}(referred to as BS), SCORE \cite{score}, AdaAD \cite{adaad}, the method in \cite{wang2023pmlr} (referred to as WANG), STL \cite{stl} and the work in \cite{salman2020} (referred to as Salman). The ResNet18  \cite{resnet18}, WRN-28-10 \cite{wrn} and EfficientNet-v2-s \cite{efficientnet} models trained on the clean training datasets are used as the base classifier $\mathbf{f}$. Further details of the experimental setup are as follows:
\begin{itemize}

\item \textbf{MNIST.} The classifier $\mathbf{f}$ is ResNet18, trained on the original MNIST training set for 25 epochs with a learning rate of 0.001, reaching 99.40\% accuracy on the original MNIST test set. All input pixels are normalized to $[0,1]$.  In PGD attacks, the number of iterations is 40 with a step size of $\epsilon/20$. For swRD, the coefficient $\beta$ is set to 5. The noise level is $\sigma=\tau=4/255$ for both pRD and swRD. The results are shown in Table \ref{tab:mnist}.

\item \textbf{CIFAR-10.} The classifier $\mathbf{f}$ is also ResNet18, trained on the original CIFAR-10 training set for 200 epochs with a learning rate of 0.01, achieving 95.16\% accuracy on the clean CIFAR-10 test set. In PGD attack, the number of iterations is 20, with a step size $\epsilon/10$. The noise level  is set to $\sigma=\tau=4/255$ for both pRD and swRD, and the coefficient $\beta=2$ in swRD. The results are presented in Table \ref{tab:cifar10}.

\item \textbf{CIFAR-100.} We select WRN-28-10 as the base model $\mathbf{f}$. Following the training steps outlined in a popular Github repository \footnote{https://github.com/meliketoy/wide-resnet.pytorch}, the classifier $\mathbf{f}$ achieves 82.05\% accuracy on the clean test set of CIFAR-100. The other experimental settings are the same as those for CIFAR-10. The results are presented in Table \ref{tab:cifar100}.

\item \textbf{ImageNet.} The base model $\mathbf{f}$ used for ImageNet is EfficientNet-v2-s. We utilize the well-trained model provided by PyTorch, which achieves 84.23\% accuracy on the clean ImageNet validation set. Each image is resized and cropped to $328 \times 328$ pixels. The noise level is set to $\sigma=\tau=4$ for both pRD and swRD, and the coefficient is $\beta=0.1$ in swRD. The results are presented in Table \ref{tab:imagenet}.
\end{itemize}

\begin{figure}[t]
\centering
\hspace{-5em}
    \begin{minipage}[t]{0.33\textwidth}
    \centering
     \begin{tikzpicture}
    \begin{axis}[
        title={(a) CIFAR10},
        xlabel={$k$}, ylabel={Accuracy},
        xlabel style={font=\tiny, yshift=0.5em},
        ylabel style={font=\tiny, xshift=0em,yshift=-0.5em},
        xmin=3, xmax=102, ymin=0, ymax=1.0,
        xtick={5,20,40,60,80,100},
        ytick={0,0.2,0.4,0.6,0.8,1.0},
        mark size=1pt,
        width=\linewidth,
        height=0.9\linewidth,   
        tick style={draw=none},
        tick label style={font=\tiny},
        title style={font=\tiny, yshift=-0.8em},
        legend style={font=\tiny}
        ]

        \addplot+[black, mark=triangle*, dash pattern=on 2pt off 1.5pt, mark options={fill=., draw=none}] coordinates {
        (5,0.7706)(10,0.8590)(20,0.8902)(30,0.8965)(40,0.9027)(50,0.9058)(60,0.9062)(70,0.9063)(80,0.9077)(90,0.9073)(100,0.9114)};
    
        \addplot+[black, mark=*, solid, mark options={fill=., draw=none}] coordinates {
        (5,0.5562)(10,0.4015)(20,0.1768)(30,0.1136)(40,0.0811)(50,0.0671)(60,0.0535)(70,0.0472)(80,0.0468)(90,0.0465)(100,0.0458)};
    
        \addplot+[orange, mark=triangle*, dash pattern=on 2pt off 1.5pt, mark options={fill=., draw=none}] coordinates {
        (5,0.5135)(10,0.6600)(20,0.7597)(30,0.8156)(40,0.8339)(50,0.8619)(60,0.8672)(70,0.8819)(80,0.8874)(90,0.8937)(100,0.9004)};

        \addplot+[orange, mark=*, solid, mark options={fill=., draw=none}] coordinates {
        (5,0.4704)(10,0.5765)(20,0.6061)(30,0.5808)(40,0.5273)(50,0.4641)(60,0.3956)(70,0.3215)(80,0.263)(90,0.1895)(100,0.1503)};
        
        \addplot+[cyan, mark=triangle*, dash pattern=on 2pt off 1.5pt, mark options={fill=., draw=none}] coordinates {
        (5,0.6439)(10,0.75)(20,0.8161)(30,0.847)(40,0.8611)(50,0.8704)(60,0.8769)(70,0.8809)(80,0.89)(90,0.8937)(100,0.8965)};
        
        \addplot+[cyan, mark=*, solid, mark options={fill=., draw=none}] coordinates {
        (5,0.6193)(10,0.7015)(20,0.7444)(30,0.7327)(40,0.7283)(50,0.7118)(60,0.6935)(70,0.6779)(80,0.6479)(90,0.63)(100,0.6044)};
        
    \end{axis}
\end{tikzpicture} 
    \end{minipage}
\hspace{-2.5em}
    \begin{minipage}[t]{0.33\textwidth}
    \centering
    \begin{tikzpicture}
    \begin{axis}[
        title={(b) CIFAR100},
        xlabel={$k$}, 
        xlabel style={font=\tiny, yshift=0.5em},
        xmin=3, xmax=102, ymin=0, ymax=0.9,
        xtick={5,20,40,60,80,100},
        ytick={0,0.2,0.4,0.6,0.8,1.0},
        mark size=1pt,
        width=\linewidth,
        height=0.9\linewidth,       
        tick style={draw=none},
        tick label style={font=\tiny},
        label style={font=\tiny},
        title style={font=\tiny, yshift=-0.8em},
        legend style={font=\tiny}
        ]

        \addplot+[black, mark=triangle*, dash pattern=on 2pt off 1.5pt, mark options={fill=., draw=none}] coordinates {
        (5,0.4980)(10,0.6243)(20,0.6911)(30,0.7101)(40,0.7163)(50,0.7221)(60,0.7214)(70,0.7260)(80,0.7234)(90,0.7259)(100,0.7252)};
    
        \addplot+[black, mark=*, solid, mark options={fill=., draw=none}] coordinates {
        (5,0.3815)(10,0.3440)(20,0.2032)(30,0.1417)(40,0.1152)(50,0.0948)(60,0.09385)(70,0.0922)(80,0.0970)(90,0.0938)(100,0.09514)};
    
        \addplot+[orange, mark=triangle*, dash pattern=on 2pt off 1.5pt, mark options={fill=., draw=none}] coordinates {
        (5,0.2784)(10,0.4215)(20,0.5396)(30,0.5986)(40,0.6375)(50,0.6631)(60,0.6758)(70,0.6921)(80,0.699)(90,0.7093)(100,0.7214)};
    
        \addplot+[orange, mark=*, solid, mark options={fill=., draw=none}] coordinates {
        (5,0.2595)(10,0.3615)(20,0.4359)(30,0.4282)(40,0.4036)(50,0.3650)(60,0.3191)(70,0.2777)(80,0.2204)(90,0.1942)(100,0.1663)};
        
        \addplot+[cyan, mark=triangle*, dash pattern=on 2pt off 1.5pt, mark options={fill=., draw=none}] coordinates {
        (5,0.3667)(10,0.4973)(20,0.5857)(30,0.6235)(40,0.6468)(50,0.6595)(60,0.6731)(70,0.6897)(80,0.6913)(90,0.6934)(100,0.7068)};
        
        \addplot+[cyan, mark=*, solid, mark options={fill=., draw=none}] coordinates {
        (5,0.3498)(10,0.4573)(20,0.5246)(30,0.5505)(40,0.5447)(50,0.5414)(60,0.5267)(70,0.5146)(80,0.5120)(90,0.4990)(100,0.4809)};
        
    \end{axis}
\end{tikzpicture}
    \end{minipage}
\hspace{-2.5em}
    \begin{minipage}[t]{0.33\textwidth}
    \centering
    \begin{tikzpicture}
    \begin{axis}[
        title={(c) ImageNet},
        xlabel={$k$},
        xlabel style={font=\tiny, yshift=0.5em},
        xmin=3, xmax=102, ymin=0, ymax=0.8,
        xtick={5,20,40,60,80,100},
        ytick={0,0.2,0.4,0.6,0.8,1.0},    
        mark size=1pt,
        width=\linewidth,
        height=0.9\linewidth,       
        tick style={draw=none},
        tick label style={font=\tiny},
        label style={font=\tiny},
        title style={font=\tiny, yshift=-0.8em},
        legend style={font=\tiny},
        legend style={
            at={(1.07,0.5)},      
            anchor=west,
            font=\tiny
        }
        ]

        \addplot+[black, mark=triangle*, dash pattern=on 2pt off 1.5pt, mark options={fill=., draw=none}] coordinates {
        (5,0.397)(10,0.568)(20,0.6490)(30,0.68)(40,0.70)(50,0.71)(60,0.728)(70,0.733)(80,0.745)(90,0.753)(100,0.765)};
        \addlegendentry{${\rm Acc}_{\rm nat}$: RandDisc}
        
        \addplot+[black, mark=*, solid, mark options={fill=., draw=none}] coordinates {
        (5,0.39)(10,0.42)(20,0.45)(30,0.47)(40,0.42)(50,0.35)(60,0.3349)(70,0.33)(80,0.323)(90,0.322)(100,0.32)};
        \addlegendentry{${\rm Acc}_{\rm rob}$: RandDisc}

        \addplot+[orange, mark=triangle*, dash pattern=on 2pt off 1.5pt, mark options={fill=., draw=none}] coordinates {
        (5,0.327)(10,0.538)(20,0.641)(30,0.65)(40,0.702)(50,0.701)(60,0.708)(70,0.725)(80,0.736)(90,0.739)(100,0.74)};
        \addlegendentry{${\rm Acc}_{\rm nat}$: pRD}
    
        \addplot+[orange, mark=*, solid, mark options={fill=., draw=none}] coordinates {
        (5,0.3)(10,0.48)(20,0.58)(30,0.62)(40,0.66)(50,0.668)(60,0.659)(70,0.654)(80,0.649)(90,0.641)(100,0.63)};
        \addlegendentry{${\rm Acc}_{\rm rob}$: pRD}
        
        \addplot+[cyan, mark=triangle*, dash pattern=on 2pt off 1.5pt, mark options={fill=., draw=none}] coordinates {
        (5,0.384)(10,0.573)(20,0.675)(30,0.693)(40,0.715)(50,0.721)(60,0.739)(70,0.741)(80,0.758)(90,0.763)(100,0.778)};
        \addlegendentry{${\rm Acc}_{\rm nat}$: swRD}
        
        \addplot+[cyan, mark=*, solid, mark options={fill=., draw=none}] coordinates {
        (5,0.341)(10,0.541)(20,0.618)(30,0.665)(40,0.675)(50,0.677)(60,0.6787)(70,0.665)(80,0.66)(90,0.654)(100,0.652)};
        \addlegendentry{${\rm Acc}_{\rm rob}$: swRD}
        
    \end{axis}
\end{tikzpicture}
    \end{minipage}
\caption{Natural and robust accuracies (y-axis) of our defenses under AutoAttack with varying $k$ values (x-axis) on three datasets. Experimental settings: $\epsilon=8/255$ for CIFAR-10 and CIFAR-100, $\epsilon =4/255$ for ImageNet; patch size $s=2$ for pRD and window size $s=3$ for swRD.}
\label{fig:all-k}
\end{figure}

\begin{figure}[t]
\centering
\hspace{-5em}
    \begin{minipage}[t]{0.33\textwidth}
    \centering
     \begin{tikzpicture}
    \begin{axis}[
        title={(a) CIFAR10},
        xlabel={$s$}, ylabel={Accuracy},
        xlabel style={font=\tiny, yshift=0.5em},
        ylabel style={font=\tiny, xshift=0em,yshift=-0.5em},
        xmin=0.8, xmax=4.2, ymin=0, ymax=1.0,
        xtick={1,2,3,4},
        ytick={0,0.2,0.4,0.6,0.8,1.0},
        mark size=1pt,
        width=\linewidth,
        height=0.9\linewidth,   
        tick style={draw=none},
        tick label style={font=\tiny},
        title style={font=\tiny, yshift=-0.8em},
        legend style={font=\tiny}
        ]

        \addplot+[orange, mark=triangle*, dash pattern=on 2pt off 1.5pt, mark options={fill=., draw=none}] coordinates {(1,0.89)(2,0.7943)(3,0.78)(4,0.805)};
        \addplot+[orange, mark=*, solid, mark options={fill=., draw=none}] coordinates {(1,0.1392)(2,0.6043)(3,0.572)(4,0.387)};
        \addplot+[cyan, mark=triangle*, dash pattern=on 2pt off 1.5pt, mark options={fill=., draw=none}] coordinates {(1,0.89)(2,0.855)(3,0.8342)(4,0.80)};
        \addplot+[cyan, mark=*, solid, mark options={fill=., draw=none}] coordinates {(1,0.1392)(2,0.65)(3,0.7367)(4,0.7231)};

    \end{axis}
\end{tikzpicture} 
    \end{minipage}
\hspace{-2em}
    \begin{minipage}[t]{0.33\textwidth}
    \centering
    \begin{tikzpicture}
    \begin{axis}[
        title={(b) CIFAR100},
        xlabel={$s$}, 
        xlabel style={font=\tiny, yshift=0.5em},
        xmin=0.8, xmax=4.2, ymin=0, ymax=1.0,
        xtick={1,2,3,4},
        ytick={0,0.2,0.4,0.6,0.8,1.0},
        mark size=1pt,
        width=\linewidth,
        height=0.9\linewidth,       
        tick style={draw=none},
        tick label style={font=\tiny},
        label style={font=\tiny},
        title style={font=\tiny, yshift=-0.8em},
        legend style={font=\tiny}
        ]

        \addplot+[orange, mark=triangle*, dash pattern=on 2pt off 1.5pt, mark options={fill=., draw=none}] coordinates {(1,0.7006)(2,0.5771)(3,0.569)(4,0.571)};
        \addplot+[orange, mark=*, solid, mark options={fill=., draw=none}] coordinates {(1,0.1652)(2,0.4364)(3,0.3998)(4,0.297)};
        \addplot+[cyan, mark=triangle*, dash pattern=on 2pt off 1.5pt, mark options={fill=., draw=none}] coordinates {(1,0.7006)(2,0.6478)(3,0.6063)(4,0.586)};
        \addplot+[cyan, mark=*, solid, mark options={fill=., draw=none}] coordinates {(1,0.1652)(2,0.52)(3,0.5378)(4,0.565)};
        
    \end{axis}
\end{tikzpicture}
    \end{minipage}
\hspace{-2em}
    \begin{minipage}[t]{0.33\textwidth}
    \centering
    \begin{tikzpicture}
    \begin{axis}[
        title={(c) ImageNet},
        xlabel={$s$},
        xlabel style={font=\tiny, yshift=0.5em},
        xmin=0.8, xmax=4.2, ymin=0, ymax=1.0,
        xtick={1,2,3,4},
        ytick={0,0.2,0.4,0.6,0.8,1.0},
        mark size=1pt,
        width=\linewidth,
        height=0.9\linewidth,       
        tick style={draw=none},
        tick label style={font=\tiny},
        label style={font=\tiny},
        title style={font=\tiny, yshift=-0.8em},
        legend style={font=\tiny},
        legend style={
            at={(1.07,0.5)},      
            anchor=west,
            font=\tiny
        }
        ]

        \addplot+[orange, mark=triangle*, dash pattern=on 2pt off 1.5pt, mark options={fill=., draw=none}] coordinates {(1,0.6769)(2,0.6556)(3,0.62)(4,0.592)};
        \addlegendentry{${\rm Acc}_{\rm nat}$: pRD}
        
        \addplot+[orange, mark=*, solid, mark options={fill=., draw=none}] coordinates {(1,0.4649)(2,0.6236)(3,0.602)(4,0.585)};
        \addlegendentry{${\rm Acc}_{\rm rob}$: pRD}
        
        \addplot+[cyan, mark=triangle*, dash pattern=on 2pt off 1.5pt, mark options={fill=., draw=none}] coordinates {(1,0.6769)(2,0.702)(3,0.6932)(4,0.662)};
        \addlegendentry{${\rm Acc}_{\rm nat}$: swRD}
        
        \addplot+[cyan, mark=*, solid, mark options={fill=., draw=none}] coordinates {(1,0.4649)(2,0.63)(3,0.675)(4,0.66)};
        \addlegendentry{${\rm Acc}_{\rm rob}$: swRD}
        
    \end{axis}
\end{tikzpicture}
    \end{minipage}
\caption{Natural and robust accuracies (y-axis) of our defenses against AutoAttack with different values of $s$ (x-axis) on three datasets. Experimental Setting: $\epsilon=8/255$; $k=25$ for CIFAR-10 and CIFAR-100; $\epsilon=4/255$; $k=30$ for ImageNet. Note: both pRD and swRD are equivalent to RandDisc when $s=1$.}
\label{fig:all-s}
\end{figure}

\begin{figure}[t]
\centering
\hspace{-5 em}
    \begin{minipage}[t]{0.43\textwidth}
    \centering
    \begin{tikzpicture}
\begin{axis}[
    title={(a) Trade-off between ${\rm Acc}_{\rm nat}$ and ${\rm Acc}_{\rm rob}$},
    xlabel={${\rm Acc}_{\rm nat}$},
    ylabel={${\rm Acc}_{\rm rob}$},
    xlabel style={font=\tiny, yshift=0.5em},
    ylabel style={font=\tiny, xshift=0em,yshift=-0.5em},
    xmin=0.18, xmax=0.7,
    ymin=0.18, ymax=0.57,
    xtick={0.2, 0.3, 0.4, 0.5, 0.6},
    ytick={0.2, 0.25, 0.3, 0.35, 0.4, 0.45, 0.5},
    width=\linewidth,
    height=\linewidth,
    tick label style={font=\tiny},
    tick style={draw=none},
    label style={font=\tiny},
    title style={font=\tiny, yshift=-0.8em},
    legend style={font=\tiny, at={(0.02,0.98)},
    anchor=north west,
    draw=none,
    fill=none,
    inner sep=1pt},
    mark size=1.5pt,
    every node near coord/.append style={font=\tiny},
    clip=false
]


\addplot+[black, mark=*, solid, mark options={fill=., draw=none}, 
    nodes near coords,
    every node near coord/.append style={font=\tiny},
    point meta=explicit symbolic]
    coordinates {
    (0.389,0.359)   [5]
    (0.545,0.443)  [10]
    (0.609,0.459)  [20]
    (0.617,0.439)   [30]
    (0.633,0.410)  [40]
    (0.6469,0.3845) [50]
};
\addlegendentry{RD}
\addplot+[orange, mark=*, solid, mark options={fill=., draw=none}, 
    nodes near coords,
    every node near coord/.append style={font=\tiny},
    point meta=explicit symbolic]
    coordinates {
        (0.21,0.20) [5]
        (0.399,0.355) [10]
        (0.474,0.439) [20] 
        (0.527,0.471) [30]
        (0.563,0.495) [40]
        (0.5856,0.5136) [50]
    };
\addlegendentry{pRD}

\addplot+[cyan, mark=*, solid, mark options={fill=., draw=none}, 
    nodes near coords,
    every node near coord/.append style={font=\tiny},
    point meta=explicit symbolic]
    coordinates {
        (0.322,0.303) [5]
        (0.508,0.443) [10]
        (0.565,0.470) [20]
        (0.598,0.480) [30]
        (0.604,0.495) [40]
        (0.632,0.522) [50]
    };
\addlegendentry{swRD}
\end{axis}
\end{tikzpicture}
    \end{minipage}
    \hspace{-2 em}
    \begin{minipage}[t]{0.43\textwidth}
    \centering
    \begin{tikzpicture}
\begin{axis}[
    title={(b) Effects of the Noise},
    xlabel={Noise level},
    ylabel={Accuracy},
    xlabel style={font=\tiny, yshift=0.5em},
    ylabel style={font=\tiny, xshift=0em,yshift=-0.5em},
    xmin=-0.7, xmax=20.5,
    ymin=0.5, ymax=0.85,
    xtick={0,2,...,20},
    ytick={0.5,0.55,...,0.9},
    tick label style={font=\scriptsize},
    tick style={draw=none},
    label style={font=\scriptsize},
    title style={font=\tiny, yshift=-0.8em},
    legend style={font=\scriptsize, at={(1.05,0.5)}, anchor=west},
    width=\linewidth,
    height=\linewidth,
    mark size=1.5pt,
    legend cell align={left}
]

\addplot+[orange, mark=triangle*, dash pattern=on 2pt off 1.5pt, mark options={fill=.}] coordinates {
(0,0.7998)(2,0.7986)(4,0.7940)(6,0.7913)(8,0.7849)(10,0.7750)
(12,0.7637)(14,0.7550)(16,0.7423)(18,0.727)(20,0.718)};
\addlegendentry{${\rm Acc}_{\rm nat}$: pRD}

\addplot+[orange, mark=square*, solid, mark options={fill=.}] coordinates {
(0,0.6013)(2,0.6040)(4,0.5954)(6,0.5995)(8,0.5880)(10,0.5741)
(12,0.5624)(14,0.5592)(16,0.5551)(18,0.5468)(20,0.54)};
\addlegendentry{${\rm Acc}_{\rm rob}$: pRD}

\addplot+[cyan, mark=triangle*, dash pattern=on 2pt off 1.5pt, mark options={fill=.}] coordinates {
(0,0.8352)(2,0.8342)(4,0.8360)(6,0.8333)(8,0.8321)(10,0.8311)
(12,0.8321)(14,0.8321)(16,0.831)(18,0.828)(20,0.827)};
\addlegendentry{${\rm Acc}_{\rm nat}$: swRD}

\addplot+[cyan, mark=square*, solid, mark options={fill=.}] coordinates {
(0,0.7356)(2,0.736)(4,0.7420)(6,0.7336)(8,0.7377)(10,0.7290)
(12,0.7260)(14,0.7192)(16,0.7034)(18,0.7014)(20,0.696)};
\addlegendentry{${\rm Acc}_{\rm rob}$: swRD}

\end{axis}
\end{tikzpicture}
    \end{minipage}
\caption{(a) Robust–natural accuracy trade-off of our methods and RandDisc on ImageNet, where numbers (i.e., $5,10,20,\dots$) indicate $k$ values. (b) Natural and robust accuracies of pRD and swRD on CIFAR-10 against AutoAttack with varying noise levels $\sigma$ and $\tau$ ($\epsilon=8/255$, $s=2/3$, $k=25$).}
\label{fig:tradeoff&noise}
\end{figure}

The experimental findings indicate that, except for AutoAttack on MNIST, the proposed pRD exhibits greater robustness compared to RandDisc, while swRD consistently outperforms all other defenses. We then study the impact of different parameters on accuracy of our defenses. First, we observe that the number $k$ of cluster centers has a significant effect on both natural and robust accuracy, as illustrated in Figure \ref{fig:all-k}. Specifically, increasing the value of $k$ leads to a rise in natural accuracy of RandDisc, pRD and swRD, but the robust accuracy of RandDisc decreases, showing a trade-off between natural and robust accuracy. However, such trade-off is not observed in either pRD or swRD, e.g., the left of Figure \ref{fig:tradeoff&noise} for CIFAR-100. In addition,  Figure \ref{fig:all-s} shows a clear trend that increasing the window size $s$ reduces the gap between natural and robust accuracies of swRD.  This agrees with the experimental observations in next subsection comparing certified robustness. Finally, as shown in the right of Figure \ref{fig:tradeoff&noise}, the noise level $\sigma$ and $\tau$ also impact the model performance. Specifically, pRD and swRD without any noise may achieve higher natural accuracy. However, introducing a suitable amount of noise can enhance the robustness of our defenses. In \ref{apdx: adding noise once}, we also provide the natural and robust accuracy with different noise levels when adding noise only once. 
\subsection{Evaluation of Certified Accuracy}
\label{sec: exp-p1}
Given a pair $(x,y)$, where $x$ is a natural image and $y$ is its true label, Proposition 1 provides an upper bound on the true robust error of $(x,y)$ with respect to a classifier $\mathbf{f}$ protected by our defenses under an attack $\mathcal{A}$. For convenience, let the right-hand side in Eq. \ref{equ:pro1} be denoted as $\gamma$. Then the certified accuracy can be expressed as 
\begin{equation}
    \text{Acc}_{\rm ctf}(x,y|\mathcal{A}, \mathbf{f},\mathbf{Q}_{\hat{X}|X})=1-\text{Err}(x,y|\mathcal{A}, \mathbf{f},\mathbf{Q}_{\hat{X}|X})\geq 1-\gamma,
\end{equation}
which provides a lower bound on the true robust accuracy of $(x,y)$ with respect to $\mathbf{f}$ protected by our defenses under $\mathcal{A}$.

\begin{figure}[t]
\centering
\hspace{-5em}
    \begin{minipage}[t]{0.33\textwidth}
    \centering
     \begin{tikzpicture}
    \begin{axis}[
        title={(a) MNIST},
        xlabel={$\epsilon$}, ylabel={${\rm Acc}_{\rm ctf}$},
        xlabel style={font=\tiny, yshift=0.5em},
        ylabel style={font=\tiny, xshift=0em,yshift=-0.5em},
        xmin=-0.02, xmax=0.42, ymin=-0.02, ymax=1.02,
        xtick={0, 0.1 , 0.2, 0.3, 0.4},
        ytick={0, 0.2,0.4,0.6,0.8,1.0},
        mark size=1pt,
        width=\linewidth,
        height=0.9\linewidth,   
        tick style={draw=none},
        tick label style={font=\tiny},
        title style={font=\tiny, yshift=-0.8em},
        legend style={font=\tiny}
        ]

        \addplot+[black, mark=*, solid, mark options={fill=., draw=none}] coordinates {
        (0.0,0.9652) (0.1,0.9502) (0.2,0.9253) (0.3,0.8761) (0.4,0.0054)};

        \addplot+[orange, mark=*, solid, mark options={fill=., draw=none}] coordinates {
        (0.0,0.9752) (0.1,0.9350) (0.2,0.8508) (0.3,0.7020) (0.4,0.4808)};

        \addplot+[cyan, mark=*, solid, mark options={fill=., draw=none}] coordinates {
        (0.0,0.9947) (0.1,0.9813) (0.2,0.9554) (0.3,0.8903) (0.4,0.5702)};

    \end{axis}
\end{tikzpicture} 
    \end{minipage}
\hspace{-2em}
    \begin{minipage}[t]{0.33\textwidth}
    \centering
    \begin{tikzpicture}
    \begin{axis}[
        title={(b) CIFAR10},
        xlabel={$\epsilon$},
        xlabel style={font=\tiny, yshift=0.5em},
        xmin=-0.2, xmax=8.2, ymin=-0.03, ymax=0.83,
        xtick={0,2,4,6,8},
        ytick={0,0.2,0.4,...,1.0},
        mark size=1pt,
        width=\linewidth,
        height=0.9\linewidth,       
        tick style={draw=none},
        tick label style={font=\tiny},
        label style={font=\tiny},
        title style={font=\tiny, yshift=-0.8em},
        ]

        \addplot+[black, mark=*, solid, mark options={fill=., draw=none}] coordinates {
        (0,0.7650) (2,0.2950) (4,0.0053) (6,0.0012) (8,0)};
        \addplot+[orange, mark=*, solid, mark options={fill=., draw=none}] coordinates {
        (0,0.6250) (2,0.4897) (4,0.3651) (6,0.3403) (8,0.1902)};
        \addplot+[cyan, mark=*, solid, mark options={fill=., draw=none}] coordinates {
        (0,0.7892) (2,0.7154) (4,0.6507) (6,0.6411) (8,0.4826)};
        
    \end{axis}
\end{tikzpicture}
    \end{minipage}
\hspace{-2em}
    \begin{minipage}[t]{0.33\textwidth}
    \centering
    \begin{tikzpicture}
    \begin{axis}[
        title={(c) CIFAR100},
        xlabel={$\epsilon$},
        xlabel style={font=\tiny, yshift=0.5em},
        xmin=-0.2, xmax=8.2, ymin=-0.03, ymax=0.63,
        xtick={0,2,4,6,8},
        ytick={0,0.1,0.2,...,1.0},
        mark size=1pt,
        width=\linewidth,
        height=0.9\linewidth,       
        tick style={draw=none},
        tick label style={font=\tiny},
        label style={font=\tiny},
        title style={font=\tiny, yshift=-0.8em},
        legend style={font=\tiny},
        legend style={
            at={(1.07,0.5)},   
            anchor=west,
            font=\tiny
        }
        ]

    \addplot+[black, mark=*, solid, mark options={fill=., draw=none}] coordinates {
    (0,0.59824) (2,0.016379) (4,0.0) (6,0.0) (8,0.0)};
    \addlegendentry{RandDisc}
    
    \addplot+[orange, mark=*, solid, mark options={fill=., draw=none}] coordinates {
    (0,0.4485) (2,0.1360) (4,0.1203) (6,0.1031) (8,0.0733)};
    \addlegendentry{pRD}
    
    \addplot+[cyan, mark=*, solid, mark options={fill=., draw=none}] coordinates {
    (0,0.5782) (2,0.5322) (4,0.3493) (6,0.3094) (8,0.1349)};
    \addlegendentry{swRD}
        
    \end{axis}
\end{tikzpicture}
    \end{minipage}
\caption{Comparison of ${\rm Acc}_{\rm ctf}$ (y-axis) between RandDisc and our defenses against AutoAttack on three datasets, where the x-axis denotes the perturbation size $\epsilon$.}
\label{fig:ctf}
\end{figure}

To evaluate the certified robustness in practice, we sample $m=1,000$ quantized images for each input and set $\delta=0.005$ to compute the right-hand side in Eq. \ref{equ:pro1}. We conduct the experiments for RandDisc, pRD and swRD on MNIST, CIFAR-10 and CIFAR-100 datasets. On MNIST, $k=2$ for all three defenses, while $s=2$ for pRD and swRD; On CIFAR-10 and CIFAR-100, $k=25$ for all three defenses, while $s=2$ for pRD and $s=3$ swRD.  We set $\mathcal{A}$ to be AutoAttack and $\mathbf{f}$ to be ResNet18 (for MNIST and CIFAR-10) and WRN-28-10 (for CIFAR-100).
The certified robustness is the average value of $\text{Acc}_{\rm ctf}(x,y|\mathcal{A}, \mathbf{f},\mathbf{Q}_{\hat{X}|X})$ on all images in the testing set. As shown in Figure \ref{fig:ctf}, our defenses, particularly, swRD, outperform RandDisc in certified robustness for almost all evaluated perturbation radii. We will observe the similar trends as the results in Section \ref{sec:exp-kld}.

\subsection{Evaluation of $U_{\rm KL}$}
\label{sec:exp-kld}
In Section \ref{sec: pro1&2}, we demonstrate that certified robustness can be evaluated by comparing the upper bounds on the KL divergence described in Equation \ref{equ:kl-c}. To facilitate this comparison, we develop a technique to numerically estimate this bound. 

For a given set $C$ of $k$ cluster centers, let vector $\overline{C}\in {\mathbb R}^{kqs^2}$ be constructed by first sorting the elements of $C$ in ascending order of their $\ell_2$-norm and then concatenating them into a vector. Let $T$ denote the mapping from $C$ to $\overline{C}$. The following can be proved.

\textbf{Lemma 1.} \textit{For a given clean image $x$ and its corresponding adversarial image $x'$ (i.e. $x'=\mathcal{A}(x)$ for any attack $\mathcal{A}$), we have}
\begin{equation}
\label{equ:kl prd}
\begin{aligned}
    U_{\rm KL}^{\rm pRD}(x) &= 
     {\rm KL}(\mathbf{Q}_{\overline{C}|X=x}^{\rm pRD}||\mathbf{Q}_{\overline{C}|X=x'}^{\rm pRD}) +\\
     &\sum_{i=1}^{n_1}\mathbb{E}_{c\sim \mathbf{Q}_{C|X=x}^{\rm pRD}}[{\rm KL}(\mathbf{Q}_{\hat{P}_i|C=c,X=x}^{\rm pRD}||\mathbf{Q}_{\hat{P}_i|C=c,X=x'}^{\rm pRD})],
\end{aligned}
\end{equation}
\begin{equation}
\label{equ:kl swrd}
\begin{aligned}
    U_{\rm KL}^{\rm swRD}(x) = 
     &{\rm KL}(\mathbf{Q}_{\overline{C}|X=x}^{\rm swRD}||\mathbf{Q}_{\overline{C}|X=x'}^{\rm swRD}) + \\
    &\sum_{i=1}^{n_2}\mathbb{E}_{c\sim \mathbf{Q}_{C|X=x}^{\rm swRD}}[{\rm KL}(\mathbf{Q}_{\hat{X}_i|C=c,X=x}^{\rm swRD}||\mathbf{Q}_{\hat{X}_i|C=c,X=x'}^{\rm swRD})],
\end{aligned}
\end{equation}
\textit{where $\hat{P}_i$ represents the $i$-th patch obtained in Eq. \ref{equ:replace_prd}, $\hat{X}_i$ represents the $i$-th pixel obtained in Eq. \ref{equ:replace_swrd}, $n_1$ is the number of patches obtained in pRD and $n_2$ is the number of pixels for a given input.}

By this lemma, the computation of KL divergence between two distributions of random sets can be approximated by its upper bound (the KL divergence between two distributions of random vectors). In practice, for each clean image $x$, we randomly draw 10,000 cluster-center sets $C$, and for each $C$, compute $\overline{C}=T(C)$. We model the distribution $\mathbf{{Q}}_{\overline{C}|X=x}$ as a Mixture of $K$ Gaussians and fit the samples $\{\overline{C}\}$ to estimate its parameters. 
We denote the estimated $\mathbf{Q}_{\overline{C}|X=x}$ as $\mathbf{\hat{Q}}_{\overline{C}|X=x}$. The same procedure is applied to the adversarial example $x'$ of $x$ to obtain an estimate $\mathbf{\hat{Q}}_{\overline{C}|X=x'}$ for the distribution $\mathbf{{Q}}_{\overline{C}|X=x'}$. Note:
\begin{equation}
\label{equ:MC}
\begin{aligned}
    {\rm KL}(\mathbf{\hat{Q}}_{\overline{C}|X=x}&||\mathbf{\hat{Q}}_{\overline{C}|X=x'}) =
    \mathbb{E}_{c\sim \mathbf{\hat{Q}}_{\overline{C}|X=x}}[\log \mathbf{\hat{Q}}_{\overline{C}|X=x}&(c)-\log \mathbf{\hat{Q}}_{\overline{C}|X=x'}(c)].
\end{aligned}
\end{equation}

This divergence can be estimated using a Monte Carlo method, where the expectation is approximated by drawing 100,000 samples from $\mathbf{\hat{Q}}_{\overline{C}|X=x}$ and taking the average. For the second term on the right-hand side of Equation \ref{equ:kl prd} and \ref{equ:kl swrd}, we use a similar method to estimate the distribution of each patch or pixel and subsequently compute the KL divergence. Notably, our experiments indicate that the magnitude of the second term is negligibly small (approximately $1e-5$) compared to the first term, rendering it inconsequential. Therefore, we focus on evaluating the KL divergence between the distributions of cluster centers. 

The estimated values of KL divergence are highly sensitive to the number $K$ of Gaussians used in Gaussian Mixture Model, leading to significantly different ranges of resulting values.  To ensure a fair comparison among pRD, swRD and RandDisc, we set $K=8$ consistently across all three methods.

\begin{figure}[t]
\centering
\hspace{-5em}
    \begin{minipage}[t]{0.33\textwidth}
    \centering
     \begin{tikzpicture}
    \begin{axis}[
        title={(a) MNIST},
        xlabel={$\epsilon$}, ylabel={KL Divergence},
        xlabel style={font=\tiny, yshift=0.5em},
        ylabel style={font=\tiny, xshift=0em,yshift=-0.5em},
        xmin=-0.03, xmax=0.35, ymin=-50, ymax=900,
        xtick={0, 0.1 , 0.2, 0.3, 0.4},
        ytick={0, 200, 400, ..., 800},
        mark size=1pt,
        width=\linewidth,
        height=0.9\linewidth,   
        tick style={draw=none},
        tick label style={font=\tiny},
        title style={font=\tiny, yshift=-0.8em},
        legend style={font=\tiny}
        ]

        \addplot+[black, mark=*, solid, mark options={fill=., draw=none}] coordinates {
        (0,0.01) (0.02,0.05) (0.04,0.12) (0.08,240) (0.12,230)
(0.16,377) (0.20,600) (0.24,620) (0.28,660) (0.32,870)};

        \addplot+[orange, mark=*, solid, mark options={fill=., draw=none}] coordinates {
        (0,0.03) (0.02,3.00) (0.04,13.2) (0.08,170) (0.12,160)
(0.16,190) (0.20,197) (0.24,210) (0.28,200) (0.32,205)};

        \addplot+[cyan, mark=*, solid, mark options={fill=., draw=none}] coordinates {
        (0,0.02) (0.02,1.80) (0.04,8.00) (0.08,20) (0.12,25)
(0.16,33) (0.20,45) (0.24,37) (0.28,60) (0.32,90)};

    \end{axis}
\end{tikzpicture} 
    \end{minipage}
\hspace{-2em}
    \begin{minipage}[t]{0.33\textwidth}
    \centering
    \begin{tikzpicture}
    \begin{axis}[
        title={(b) CIFAR10},
        xlabel={$\epsilon$},
        xlabel style={font=\tiny, yshift=0.5em},
        xmin=-1, xmax=17, ymin=-50, ymax=1150,
        xtick={0, 4, 8, ...,16},
        ytick={0, 200, 400, ..., 1000},
        mark size=1pt,
        width=\linewidth,
        height=0.9\linewidth,       
        tick style={draw=none},
        tick label style={font=\tiny},
        label style={font=\tiny},
        title style={font=\tiny, yshift=-0.8em},
        ]

        \addplot+[black, mark=*, solid, mark options={fill=., draw=none}] coordinates {
        (0,3) (2,13) (4,48) (6,110) (8,180) (10,460) (12,470) (14,960) (16,1072)};
        \addplot+[orange, mark=*, solid, mark options={fill=., draw=none}] coordinates {
        (0,11) (2,18) (4,46) (6,90) (8,125) (10,195) (12,227) (14,186) (16,200)};
        \addplot+[cyan, mark=*, solid, mark options={fill=., draw=none}] coordinates {
        (0,9) (2,11) (4,44.5) (6,80) (8,92) (10,150) (12,170) (14,180) (16,210)};
        
    \end{axis}
\end{tikzpicture}
    \end{minipage}
\hspace{-2em}
    \begin{minipage}[t]{0.33\textwidth}
    \centering
    \begin{tikzpicture}
    \begin{axis}[
        title={(c) ImageNet},
        xlabel={$\epsilon$},
        xlabel style={font=\tiny, yshift=0.5em},
        xmin=-0.5, xmax=6.5, ymin=-50, ymax=3200,
        xtick={0, 1,2,3,...,6},
        ytick={0, 500, 1000, 1500, ..., 3000},
        mark size=1pt,
        width=\linewidth,
        height=0.9\linewidth,       
        tick style={draw=none},
        tick label style={font=\tiny},
        label style={font=\tiny},
        title style={font=\tiny, yshift=-0.8em},
        legend style={font=\tiny},
        legend style={
            at={(1.07,0.5)},      
            anchor=west,
            font=\tiny
        }
        ]

    \addplot+[black, mark=*, solid, mark options={fill=., draw=none}] coordinates {
    (0,30) (1,70) (2,90) (3,788) (4,1126) (5,2344) (6,3078)};
    \addlegendentry{RandDisc}
    
    \addplot+[orange, mark=*, solid, mark options={fill=., draw=none}] coordinates {
    (0,65) (1,94) (2,122) (3,562) (4,756) (5,1187) (6,1432)};
    \addlegendentry{pRD}
    
    \addplot+[cyan, mark=*, solid, mark options={fill=., draw=none}] coordinates {
    (0,45) (1,68) (2,80) (3,288) (4,599) (5,783) (6,900)};
    \addlegendentry{swRD}
        
    \end{axis}
\end{tikzpicture}
    \end{minipage}
\caption{Comparison of ${\rm KL}(\mathbf{\hat{Q}}_{\overline{C}|X=x}||\mathbf{\hat{Q}}_{\overline{C}|X=x'})$ (y-axis) between RandDisc and our defenses against AutoAttack on three datasets, where the x-axis is the perturbation size $\epsilon$.}
\label{fig:kld}
\end{figure}

\begin{figure}[t]
\centering
\hspace{-5em}
    \begin{minipage}[t]{0.33\textwidth}
    \centering
     \begin{tikzpicture}
    \begin{axis}[
        title={(a) CIFAR10},
        xlabel={$k$}, ylabel={Gap value},
        xlabel style={font=\tiny, yshift=0.5em},
        ylabel style={font=\tiny, xshift=0em,yshift=-0.5em},
        xmin=0, xmax=55, ymin=-0.05, ymax=0.82,
        xtick={5, 10, 20, 30, 40, 50},
        ytick={0, 0.1, 0.2, ..., 0.8},
        mark size=1pt,
        width=\linewidth,
        height=0.9\linewidth,   
        tick style={draw=none},
        tick label style={font=\tiny},
        title style={font=\tiny, yshift=-0.8em},
        legend style={font=\tiny}
        ]

        \addplot+[black, mark=*, solid, mark options={fill=., draw=none}] coordinates {
        (5,0.2132) (10,0.4652) (20,0.6161) (30,0.7072) (40,0.7630) (50,0.7889)};

        \addplot+[orange, mark=*, solid, mark options={fill=., draw=none}] coordinates {
        (5,0.0442) (10,0.0785) (20,0.1331) (30,0.1590) (40,0.1913) (50,0.2373)};

        \addplot+[cyan, mark=*, solid, mark options={fill=., draw=none}] coordinates {
        (5,0.0148) (10,0.0635) (20,0.0731) (30,0.0846) (40,0.1005) (50,0.1155)};

    \end{axis}
\end{tikzpicture} 
    \end{minipage}
\hspace{-2em}
    \begin{minipage}[t]{0.33\textwidth}
    \centering
     \begin{tikzpicture}
    \begin{axis}[
        title={(b) CIFAR100},
        xlabel={$k$},
        xlabel style={font=\tiny, yshift=0.5em},
        xmin=0, xmax=55, ymin=-0.05, ymax=0.6,
        xtick={5, 10, 20, 30, 40, 50},
        ytick={0, 0.1, 0.2, ..., 0.5},
        mark size=1pt,
        width=\linewidth,
        height=0.9\linewidth,   
        tick style={draw=none},
        tick label style={font=\tiny},
        title style={font=\tiny, yshift=-0.8em},
        legend style={font=\tiny}
        ]

        \addplot+[black, mark=*, solid, mark options={fill=., draw=none}] coordinates {
        (5,0.1294) (10,0.2795) (20,0.4216) (30,0.4958) (40,0.5260) (50,0.5651)};

        \addplot+[orange, mark=*, solid, mark options={fill=., draw=none}] coordinates {
        (5,0.0218) (10,0.0661) (20,0.0789) (30,0.1128) (40,0.1487) (50,0.1729)};

        \addplot+[cyan, mark=*, solid, mark options={fill=., draw=none}] coordinates {
        (5,0.0203) (10,0.0301) (20,0.0451) (30,0.0441) (40,0.0702) (50,0.0873)};

    \end{axis}
\end{tikzpicture} 
    \end{minipage}
\hspace{-2em}
    \begin{minipage}[t]{0.33\textwidth}
    \centering
    \begin{tikzpicture}
    \begin{axis}[
        title={(c) ImageNet},
        xlabel={$k$},
        xlabel style={font=\tiny, yshift=0.5em},
        xmin=0, xmax=55, ymin=-0.05, ymax=0.42,
        xtick={5, 10, 20, 30, 40, 50},
        ytick={0, 0.1, 0.2, 0.3, 0.4},
        mark size=1pt,
        width=\linewidth,
        height=0.9\linewidth,       
        tick style={draw=none},
        tick label style={font=\tiny},
        label style={font=\tiny},
        title style={font=\tiny, yshift=-0.8em},
        legend style={font=\tiny},
        legend style={
            at={(1.07,0.5)}, 
            anchor=west,
            font=\tiny
        }
        ]

    \addplot+[black, mark=*, solid, mark options={fill=., draw=none}] coordinates {
    (5,0.0070) (10,0.1480) (20,0.1990) (30,0.2100) (40,0.2800) (50,0.3600)};
    \addlegendentry{$\Delta_{\rm RandDisc}$}
    
    \addplot+[orange, mark=*, solid, mark options={fill=., draw=none}] coordinates {
    (5,0.0270) (10,0.0580) (20,0.0610) (30,0.0300) (40,0.0420) (50,0.0330)};
    \addlegendentry{$\Delta_{\rm pRD}$}
    
    \addplot+[cyan, mark=*, solid, mark options={fill=., draw=none}] coordinates {
    (5,0.0430) (10,0.0320) (20,0.0570) (30,0.0280) (40,0.0400) (50,0.0440)};
    \addlegendentry{$\Delta_{\rm swRD}$}
        
    \end{axis}
\end{tikzpicture}
    \end{minipage}
\\
\hspace{-5em}
    \begin{minipage}[t]{0.33\textwidth}
    \centering
     \begin{tikzpicture}
    \begin{axis}[
        title={(a) CIFAR10},
        xlabel={$s$}, ylabel={Gap value},
        xlabel style={font=\tiny, yshift=0.5em},
        ylabel style={font=\tiny, xshift=0em,yshift=-0.5em},
        xmin=0.5, xmax=4.2, ymin=-0.05, ymax=0.8,
        xtick={1,2,3,4},
        ytick={0, 0.1, 0.2, ..., 0.7},
        mark size=1pt,
        width=\linewidth,
        height=0.9\linewidth,   
        tick style={draw=none},
        tick label style={font=\tiny},
        title style={font=\tiny, yshift=-0.8em},
        legend style={font=\tiny}
        ]

        
        \addplot+[orange, mark=*, solid, mark options={fill=., draw=none}] coordinates {
        (1,0.7508) (2,0.1900) (3,0.2080) (4,0.4180)};

        \addplot+[cyan, mark=*, solid, mark options={fill=., draw=none}] coordinates {
        (1,0.7508) (2,0.2050) (3,0.0975) (4,0.0769)};

    \end{axis}
\end{tikzpicture} 
    \end{minipage}
\hspace{-2em}
    \begin{minipage}[t]{0.33\textwidth}
    \centering
     \begin{tikzpicture}
    \begin{axis}[
        title={(b) CIFAR100},
        xlabel={$s$},
        xlabel style={font=\tiny, yshift=0.5em},
        xmin=0.5, xmax=4.2, ymin=-0.05, ymax=0.58,
        xtick={1,2,3,4},
        ytick={0, 0.1, 0.2, ..., 0.5},
        mark size=1pt,
        width=\linewidth,
        height=0.9\linewidth,   
        tick style={draw=none},
        tick label style={font=\tiny},
        title style={font=\tiny, yshift=-0.8em},
        legend style={font=\tiny}
        ]

        
        \addplot+[orange, mark=*, solid, mark options={fill=., draw=none}] coordinates {
        (1,0.5354) (2,0.1407) (3,0.1692) (4,0.2740)};

        \addplot+[cyan, mark=*, solid, mark options={fill=., draw=none}] coordinates {
        (1,0.5354) (2,0.1278) (3,0.0685) (4,0.0210)};

    \end{axis}
\end{tikzpicture} 
    \end{minipage}
\hspace{-2em}
    \begin{minipage}[t]{0.33\textwidth}
    \centering
    \begin{tikzpicture}
    \begin{axis}[
        title={(c) ImageNet},
        xlabel={$s$},
        xlabel style={font=\tiny, yshift=0.5em},
        xmin=0.5, xmax=4.2, ymin=-0.01, ymax=0.2355,
        xtick={1,2,3,4},
        ytick={0, 0.05, 0.10, 0.15, 0.20},
        y tick label style={
            /pgf/number format/fixed,
            /pgf/number format/precision=2
          },
        mark size=1pt,
        width=\linewidth,
        height=0.9\linewidth,       
        tick style={draw=none},
        tick label style={font=\tiny},
        label style={font=\tiny},
        title style={font=\tiny, yshift=-0.8em},
        legend style={font=\tiny},
        legend style={
            at={(1.07,0.5)},      
            anchor=west,
            font=\tiny
        }
        ]

    
    \addplot+[orange, mark=*, solid, mark options={fill=., draw=none}] coordinates {
    (1,0.2120) (2,0.0320) (3,0.0180) (4,0.0070)};
    \addlegendentry{$\Delta_{\rm pRD}$}
    
    \addplot+[cyan, mark=*, solid, mark options={fill=., draw=none}] coordinates {
    (1,0.2120) (2,0.0720) (3,0.0182) (4,0.0020)};
    \addlegendentry{$\Delta_{\rm swRD}$}
        
    \end{axis}
\end{tikzpicture}
    \end{minipage}
\caption{Gaps between natural and robust accuracies under different hyper-parameter settings (upper plots: $k$; lower plots: $s$). Results are derived from Figure~\ref{fig:all-k} and Figure~\ref{fig:all-s}. Note that $\Delta_{\rm RandDisc}$ corresponds to $\Delta_{\rm pRD}$ or $\Delta_{\rm swRD}$ when $s=1$ (i.e., the black curves in the upper plots and the left-most points in the lower plots). Evidently, $\Delta_{\rm RandDisc}$ consistently exhibits higher values than both $\Delta_{\rm pRD}$ and $\Delta_{\rm swRD}$ for $s > 1$.}
\label{fig:diff}
\end{figure}

The results for estimating KL bounds are depicted in Figure \ref{fig:kld}. For small perturbation sizes, RandDisc exhibits a lower KL divergence. However, as the perturbation radius $\epsilon$ increases, the KL divergences ${\rm KL}(\mathbf{\hat{Q}}_{\overline{C}|X=x}||\mathbf{\hat{Q}}_{\overline{C}|X=x'})$ for pRD and swRD become much lower than that of RandDisc across all datasets. This observation suggests that RandDisc may achieve stronger certified robustness than our proposed defenses under small or zero perturbations. However, as the perturbation level increases, the certified robustness of RandDisc decreases substantially and falls well below that of our defenses for large $\epsilon$. 

As noted in the previous section, the gap $\Delta: = {\rm Acc}_{\rm nat}-{\rm Acc}_{\rm rob}$ (i.e., the difference between natural and robust accuracies) exhibits a similar trend. To see this more clearly, Figure \ref{fig:diff} presents that gaps associated with our defenses across diverse datasets and hyper-parameter settings always manifest smaller values compared to those of RandDisc. Indeed, the gap $\Delta$ may be viewed as a proxy for the divergence ${\rm KL}(\mathbf{Q}_{\hat{X}|X=x}||\mathbf{Q}_{\hat{X}|X=x'})$. 
\subsection{Denoising Effects}

\begin{figure}[t]
  \centering
  \includegraphics[width=0.9\linewidth]{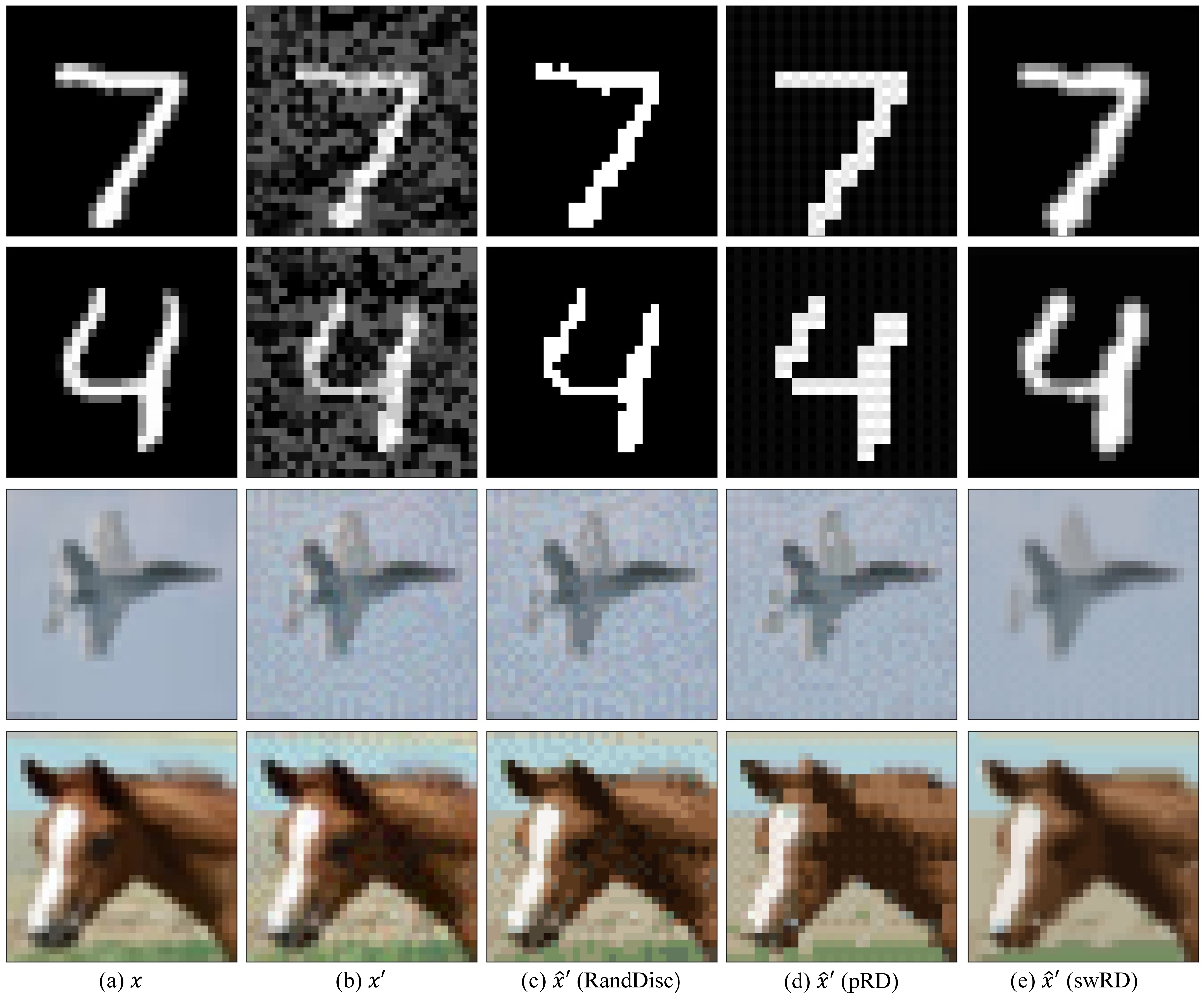}
\caption{(a) Original images $x$; (b) corresponding adversarial examples $x'$; (c) quantized versions $\hat{x}'$ by RandDisc ($k=2$ on MNIST, $k=25$ on CIFAR-10); (d) quantized versions $\hat{x}'$ by pRD ($s=2$, $k=2$ on MNIST; $s=2$, $k=25$ on CIFAR-10); (e) quantized versions $\hat{x}'$ by swRD ($s=2$, $k=2$ on MNIST; $s=3$, $k=25$ on CIFAR-10).}
\label{fig:visual} 
\end{figure}

We now perform visual inspection of the quantized images by our defenses and by RandDisc. Figure \ref{fig:visual} presents several examples. The adversarial examples $x'$ are generated by AutoAttack with $\epsilon=0.3$ on MNIST and $\epsilon=8/255$ on CIFAR-10. It is worth noting that in these examples, the adversarial perturbation primarily lies in the background of the images (which is better revealed when zooming in on the image). It can be seen that the clean images $x$ and the quantized adversarial images $\widehat{x'}$ by our defenses (especially  swRD), are much closer than those for RandDisc. Specifically, the background noise introduced by adversarial perturbation is much better removed in pRD and swRD than in RandDisc. This demonstrates a remarkable denoising effects of the proposed defenses. Notably,  unlike the defensive denoisers in previous works \cite{xie2019feature, liao2018defense, naseer2020self, jia2019comdefend}, the denoising performance of swRD does not rely on having access to the original image $x$, nor does it require training an extra network to learn the relationship between $x$ and $x'$; the sole input of swRD is the perturbed image $x'$. Its simplicity and 
its remarkable performance makes swRD much more appealing than other defensive denoisers.

\section{Experiments: Attacks on the Entire Network That Include the Defense}
\label{sec: obfuscated}
\subsection{Obfuscated Gradients Problem}
One might naturally wonder: what if the attack is specifically designed to the target network shielded with our defenses, i.e.,  the entire network depicted within the red dashed box in Figure \ref{fig:diagram}? It is straight-forward to verify that if such attacks are based on the gradient signals of the overall network, the attacks must fail. This is because in our defense schemes, the gradient signal with respect to the input is either $0$ or non-existent.

However, Athalye et al. \cite{Obfuscated} suggested that adversarial defenses that solely rely on ``obfuscated gradients" are not truly secure and that they are vulnerable to some attacks designed specifically to break defenses that obfuscating gradients. The relevance of these results to our proposed defenses lies in the fact that our defenses also have the obfuscated gradient property. Specifically, Athalye et al. \cite{Obfuscated} identified three classes of defenses with  obfuscated gradients, and our defenses belong to two of the three classes. These two classes are termed ``shattered gradients" and ``stochastic gradients" in \cite{Obfuscated}, where the former arises when the defense is non-differentiable and the latter results from randomization. It is then evident that both pRD and swRD belong to defenses with obfuscated gradients. A question naturally arises: will our defenses survive the attacks specifically designed to break such defenses?

To answer this question, we conduct experiments in Section \ref{sec: exp-ste&eot} using the two schemes in \cite{Obfuscated} designed to attack defenses in the ``shattered gradients" and ``stochastic gradients" classes. Our results show that our defenses remain effective. The key take-away from our experiments is that pRD and swRD do not simply rely on obfuscated gradients as the main defense mechanism and that their power lies in properly exploiting the input data structure via more effective quantization.

For the sake of completeness, we next provide a concise review of  the attacks designed in \cite{Obfuscated} to break defenses with ``shattered gradients" and ``stochastic gradients", namely, STE (Straight-Through Estimator) and EOT (Expectation over Transformation). Notably these methods are also gradient-based attacks, and  we only explain how they extract gradient signals.

\subsection{Gradient Computation in STE and EOT}

{\bf STE for Shattered Gradients} Suppose that the preprocessor $\mathbf{Q}_{\hat{X}|X}$ is a deterministic non-differentiable function $g:\mathcal{X}\rightarrow\mathcal{X}$. For any input $x$, in order to preserve the predicted label, $g(x)\approx x$ is usually satisfied. Then one may approximate the gradient of $g$ by $\triangledown_x g(x)\approx 1$. Then when attacking the entire model, one can approximate the gradient signal of ${\bf f}\circ g$ with respect to input, say $x_0$, by
\begin{equation}
\triangledown_x {\bf f}\circ g (x)\big|_{x=x_0}\approx\triangledown_x {\bf f}(x)\big|_{x=g(x_0)}.
\end{equation}

{\bf EOT for  Stochastic Gradients} When the preprocessor $Q_{\hat{X}|X}$ implements a stochastic mapping, we may alternatively regard the preprocessor as operating by first generating a random function $g$ and then outputting $g(x)$ for each input  $x$, and likewise $g(x)\approx x$. Then
EOT addresses stochastic gradient by computing the gradient of the model output expected over $g$. More precisely,  for an input $x_0$,  EOT computes the gradient by
\begin{equation}
\begin{aligned}
\triangledown_x \mathbb{E}_g\left[{\bf f}\circ g (x)\right]\big|_{x=x_0}&=\mathbb{E}_g\left[\triangledown_x {\bf f}\circ g (x)\big|_{x=x_0}\right]\\
&\approx\mathbb{E}_g\left[\triangledown_x {\bf f}(x)\big|_{x=g(x_0)}\right]\\
&=\mathbb{E}_{\hat{x}\sim Q_{\hat{X}|X=x_0}}\left[\triangledown_x {\bf f}(x)\big|_{x=\hat{x}}\right].
\end{aligned}
\end{equation}


\subsection{Robust Accuracy against STE and EOT}
\label{sec: exp-ste&eot}
\begin{table}
\caption{The robust accuracy (\%) of RandDisc and our defenses against the attack algorithms of PGD and AutoAttack combined with STE and EOT on three datasets. The perturbation size $\epsilon=8/255$ on both CIFAR10 and CIFAR100, while $\epsilon=4/255$ on ImageNet. The details of parameter settings are provided in Section \ref{sec: exp-ste&eot}.}
\begin{center}
\begin{tabular}{l|c|c|c|c} 
\hline
\textbf{Dataset} & \textbf{Technique in} \cite{Obfuscated} & \textbf{Attack} & \textbf{Defense} & $\mathbf{Acc_{rob}}$ \\
\hline\hline
\multirow{12}*{CIFAR10}
  & \multirow{6}*{STE}   & \multirow{3}*{PGD} & RandDisc & 27.19(0.08)\\ \cline{4-5} 
~ & ~ & ~ & pRD  & 61.21(0.08)  \\ \cline{4-5} 
~ & ~ & ~ & swRD  & 68.42(0.02)  \\ \cline{3-5} 
~ & ~ & \multirow{3}*{AutoAttack} & RandDisc & 28.39(0.11)\\ \cline{4-5} 
~ & ~ & ~ & pRD  & 67.09(0.07)   \\ \cline{4-5} 
~ & ~ & ~ & swRD  & 68.88(0.11)  \\ \cline{2-5} 

~ & \multirow{6}*{EOT}   & \multirow{3}*{PGD} & RandDisc & 27.32(0.09)\\ \cline{4-5} 
~ & ~ & ~ & pRD  & 62.51(0.08)  \\ \cline{4-5} 
~ & ~ & ~ & swRD  & 70.33(0.05)  \\ \cline{3-5} 
~ & ~ & \multirow{3}*{AutoAttack} & RandDisc & 27.87(0.13)\\ \cline{4-5} 
~ & ~ & ~ & pRD  & 67.26(0.13)   \\ \cline{4-5} 
~ & ~ & ~ & swRD  & 71.44(0.04)  \\ \hline

\multirow{12}*{CIFAR100}
  & \multirow{6}*{STE}   & \multirow{3}*{PGD} & RandDisc & 17.58(0.09)\\ \cline{4-5} 
~ & ~ & ~ & pRD  & 39.02(0.07)  \\ \cline{4-5} 
~ & ~ & ~ & swRD  & 43.07(0.11)  \\ \cline{3-5} 
~ & ~ & \multirow{3}*{AutoAttack} & RandDisc & 27.41(0.13)\\ \cline{4-5} 
~ & ~ & ~ & pRD  & 49.21(0.14)   \\ \cline{4-5} 
~ & ~ & ~ & swRD  & 48.28(0.06)  \\ \cline{2-5} 

~ & \multirow{6}*{EOT}   & \multirow{3}*{PGD} & RandDisc & 16.62(0.16)\\ \cline{4-5} 
~ & ~ & ~ & pRD  & 39.88(0.10)  \\ \cline{4-5} 
~ & ~ & ~ & swRD  & 49.05(0.13)  \\ \cline{3-5} 
~ & ~ & \multirow{3}*{AutoAttack} & RandDisc & 25.23(0.12)\\ \cline{4-5} 
~ & ~ & ~ & pRD  & 49.31(0.06)   \\ \cline{4-5} 
~ & ~ & ~ & swRD  & 50.78(0.02)  \\ \hline

\multirow{6}*{ImageNet}
  & \multirow{3}*{STE}   & \multirow{3}*{AutoAttack} & RandDisc & 39.22(0.95)\\ \cline{4-5} 
~ & ~ & ~ & pRD  & 67.12(1.04)\\ \cline{4-5} 
~ & ~ & ~ & swRD  & 64.39(0.58)\\ \cline{2-5} 

~ & \multirow{3}*{EOT}   & \multirow{3}*{AutoAttack} & RandDisc & 38.94(0.68) \\ \cline{4-5} 
~ & ~ & ~ & pRD  & 66.63(1.02)\\ \cline{4-5} 
~ & ~ & ~ & swRD  & 63.27(0.86)\\ \hline
\end{tabular}
\end{center}
\label{tab:robust-acc-obfuscated}
\end{table}
We conducted experiments to evaluate the robust accuracy of our defenses against STE and EOT, as introduced in Section \ref{sec: obfuscated}. The parameter settings for RandDisc were as follows: $k=25$ on both CIFAR10 and CIFAR100, while $k=30$ on ImageNet. For pRD, $s=2, k=25$ on both CIFAR10 and CIFAR100, and $s=2, k=30$  on ImageNet. Finally, as for swRD,  $s=2, k=15$ on CIFAR10, $s=2, k=20$ on CIFAR100, and $s=2, k=30$  on ImageNet. 

The experimental results, summarized in Table \ref{tab:robust-acc-obfuscated}, indicate that the robust accuracy of our methods slightly decreases when STE and EOT are employed. However, unlike the defenses mentioned in \cite{Obfuscated} (e.g. the pre-processor in \cite{Guo-Obfuscated-preprocessor} and the randomized method in \cite{Xie-Obfuscated-randomization}) that completely fail against STE and EOT, our defenses maintain satisfactory robustness.

These experiments confirm that although pRD and swRD are defense that obfuscate gradients, the effectiveness of these defenses has little to do with this property. Rather, their power lies in more effective exploitation of the data structure, principled by rate-distortion theory.


\section{Conclusion and Limitation}

This paper presents a vector quantization framework for preprocessing-based adversarial defenses and proposes two defenses within this framework.
We show that such defenses have certifiable robust accuracy and demonstrate the state-of-the-art performance.

Despite the power of this framework demonstrated by the two proposed defenses, we note that these defenses still have considerable room for improvement. For example, readers may observe that the natural accuracies of our defenses are not the highest among all compared approaches. This could be due to the fact that our defenses are ignorant of the input distribution and downstream classifier to be protected. Additional performance gains, particularly in natural accuracy, might be anticipated if the quantizers in these defenses are fine-tuned (possibly together with the re-training of the downstream classifier in case that option exists) using a training set. 

\clearpage
\bibliographystyle{abbrv}
\bibliography{ref}
\clearpage
\appendix
\section{Proof of Proposition 1}
\label{apdx: proof P1}
Let $\widehat{x'}$ denote the quantized version of $\mathcal{A}(x)$. Then, the relationship between the classification margin and $\ell_1$ norm can be expressed as:
\begin{equation}
\begin{aligned}
    |\rho(x,y,\mathbf{f})-\rho(\widehat{x'},y,\mathbf{f})| \leq  ||\mathbf{f}(x)-\mathbf{f}(\hat{x})||_{1} \\
    \rho(\widehat{x'},y,\mathbf{f}) \geq \rho(x,y,\mathbf{f})-  ||\mathbf{f}(x)-\mathbf{f}(\hat{x})||_{1}
\label{equ: margin_norm}
\end{aligned}
\end{equation}
By the definition of true robust error, we have 
\begin{equation}
\begin{aligned}
\text{Err}(x,y|\mathcal{A},\mathbf{f},\mathbf{Q}_{\cdot|\cdot})
&=\Pr[\rho(\widehat{x'},y,\mathbf{f}) \leq 0] \\
&\leq \Pr[\rho(x,y,\mathbf{f})-||\mathbf{f}(x)-\mathbf{f}(\widehat{x'})||_{1} \leq 0]\\
&=\Pr[||\mathbf{f}(x)-\mathbf{f}(\widehat{x'})||_1 \geq \rho(x,y,\mathbf{f})]\\
&\leq \frac{\mathbb{E}[||\mathbf{f}(x)-\mathbf{f}(\widehat{x'})||_{1}]}{\rho(x,y,\mathbf{f})}\\
\end{aligned}
\end{equation}
where the last step is obtained by Markov inequality. Let $Y_i=||\mathbf{f}(x)-\mathbf{f}(\widehat{x'}_{i})||_1$ where $\widehat{x'}_{i}$ are i.i.d drawn from $\mathbf{Q}_{\hat{X}|X=\mathcal{A}(x)}$ for $i=1,\dots,m$. Then,
\begin{equation}
\begin{aligned}
\Pr\left[\frac{1}{m}\sum_{i=1}^m Y_i-\mathbb{E}[Y_i]\leq -\beta\right ]&\leq \exp(- m\beta^2/2)\\
\Pr\left[\mathbb{E}[Y_i]\leq \frac{1}{m}\sum_{i=1}^m Y_i+\beta\right ]&\geq 1- \exp(- m\beta^2/2)\\
\end{aligned}
\end{equation}
The result of Proposition 1 is obtained by setting $\delta = \exp(- m\beta^2/2)$. $\hfill\square$

\section{Adding noise once}
\label{apdx: adding noise once}

\begin{figure}[htbp]
\centering
\begin{tikzpicture}
\begin{axis}[
    xlabel={Noise level},
    ylabel={Accuracy},
    xmin=-0.5, xmax=20.5,
    ymin=0.5, ymax=0.87,
    xtick={0,2,...,20},
    ytick={0.5,0.55,...,0.9},
    tick label style={font=\scriptsize},
    tick style={draw=none},
    label style={font=\scriptsize},
    legend style={font=\scriptsize, at={(1.05,0.5)}, anchor=west},
    width=0.5\linewidth,
    height=0.5\linewidth,
    mark size=1.5pt,
    legend cell align={left}
]

\addplot+[orange, mark=triangle*, dash pattern=on 2pt off 1.5pt, mark options={fill=orange}] coordinates {
(0,0.7998)(2,0.799)(4,0.7915)(6,0.7913)(8,0.7789)(10,0.7728)
(12,0.7515)(14,0.7389)(16,0.7321)(18,0.7101)(20,0.7095)};
\addlegendentry{${\rm Acc}_{\rm nat}$: pRD}

\addplot+[orange, mark=square*, solid, mark options={fill=.}] coordinates {
(0,0.6013)(2,0.6027)(4,0.6043)(6,0.5914)(8,0.5880)(10,0.5815)
(12,0.5814)(14,0.5768)(16,0.571)(18,0.5689)(20,0.5622)};
\addlegendentry{${\rm Acc}_{\rm rob}$: pRD}

\addplot+[cyan, mark=triangle*, dash pattern=on 2pt off 1.5pt, mark options={fill=.}] coordinates {
(0,0.8352)(2,0.8353)(4,0.8359)(6,0.8333)(8,0.8321)(10,0.8311)
(12,0.8287)(14,0.8221)(16,0.8182)(18,0.8131)(20,0.810)};
\addlegendentry{${\rm Acc}_{\rm nat}$: swRD}
\addplot+[cyan, mark=square*, solid, mark options={fill=.}] coordinates {
(0,0.7356)(2,0.7426)(4,0.7467)(6,0.7367)(8,0.7377)(10,0.7313)
(12,0.7220)(14,0.7188)(16,0.7184)(18,0.7114)(20,0.7105)};
\addlegendentry{${\rm Acc}_{\rm rob}$: swRD}

\end{axis}
\end{tikzpicture}
\caption{Natural and robust accuracies of our defenses against AutoAttack with varying noise levels $\sigma$ on CIFAR-10. The evaluation is conducted with $\tau = 0$ and $\epsilon = 8/255$; $s = 2$, $k = 25$ for pRD and $s = 3$, $k = 25$ for swRD.}
\label{fig:noise-once}
\end{figure}

Here we show the performance of our defenses when adding the isotropic Gaussian noise once, that is, by setting or $\tau=0$ in pRD and swRD. We find that the difference of adding noise once and adding noise twice is not significant, as shown in Figure \ref{fig:noise-once}. Adding noise twice ((b) in Figure \ref{fig:tradeoff&noise}) shows a slight increase in robustness, while adding noise once results in slightly higher clean accuracy. This is reasonable because adding noise twice introduces stronger randomness.

\end{document}